\documentclass{article}

\usepackage[nonatbib,final]{neurips_2020}

\usepackage[utf8]{inputenc} %
\usepackage[T1]{fontenc}    %
\usepackage{hyperref}       %
\usepackage{url}            %
\usepackage{booktabs}       %
\usepackage{amsfonts}       %
\usepackage{nicefrac}       %
\usepackage{microtype}      %

\usepackage[numbers]{natbib}
\usepackage{amsmath}
\usepackage{amssymb}
\usepackage{subfig}
\usepackage{enumitem}
\usepackage{float}
\usepackage{algorithm}
\usepackage[noend]{algorithmic}
\usepackage{graphicx}
\usepackage{bbm}
\usepackage{multirow}
\usepackage{multicol}
\usepackage{caption}
\usepackage{multicol}
\usepackage{xcolor}
\usepackage{fancyvrb}

\usepackage{varioref}
\labelformat{equation}{(#1)}

\newcommand{\e}{\epsilon}
\newcommand{\y}{\gamma}
\newcommand{\al}{\alpha}

\newcommand{\ta}{\theta}

\newcommand{\g}{\nabla}
\newcommand{\E}{\mathbb{E}}
\newcommand{\N}{\mathcal{N}}
\newcommand{\lb}{\left [}
\newcommand{\rb}{\right ]}
\newcommand{\lp}{\left (}
\newcommand{\rp}{\right )}

\newcommand{\B}{\mathcal{B}}
\newcommand{\Loss}{\mathcal{L}}

\newtheorem{theorem}{Theorem}
\newtheorem{corollary}{Corollary}
\newtheorem{observation}{Observation}

\makeatletter
\let\ORG@Gscale@box\Gscale@box
\long\def\Gscale@box#1{%
  \xdef\thelastscalefactor{#1}%
  \ORG@Gscale@box{#1}}
\makeatother

\title{An Equivalence between Loss Functions and Non-Uniform Sampling in Experience Replay}

\author{
Scott Fujimoto, David Meger, Doina Precup \\
Mila, McGill University \\
\texttt{scott.fujimoto@mail.mcgill.ca}
}

\begin{document}

\maketitle

\begin{abstract}
Prioritized Experience Replay (PER) is a deep reinforcement learning technique in which agents learn from transitions sampled with non-uniform probability proportionate to their temporal-difference error. We show that any loss function evaluated with non-uniformly sampled data can be transformed into another uniformly sampled loss function with the same expected gradient. Surprisingly, we find in some environments PER can be replaced entirely by this new loss function without impact to empirical performance. Furthermore, this relationship suggests a new branch of improvements to PER by correcting its uniformly sampled loss function equivalent. We demonstrate the effectiveness of our proposed modifications to PER and the equivalent loss function in several MuJoCo and Atari environments. %
\end{abstract}

\section{Introduction}

The use of non-uniform sampling in deep reinforcement learning originates from a technique known as prioritized experience replay (PER) \cite{PrioritizedExpReplay}, in which high error transitions are sampled with increased probability, enabling faster propagation of rewards and accelerating learning by focusing on the most critical data. An ablation study over the many improvements to deep Q-learning \cite{hessel2017rainbow} found PER to be the most critical extension for overall performance. 
However, while the motivation for PER is intuitive, this commonly used technique lacks a critical theoretical foundation. In this paper, we develop analysis enabling us to understand the benefits of non-uniform sampling and suggest modifications to PER that both simplify and improve the empirical performance of the algorithm. 

In deep reinforcement learning, value functions are approximated with deep neural networks, trained with a loss over the temporal-difference error of previously seen transitions. The expectation of this loss, over the sampling distribution of transitions, determines the effective gradient which is used to find minima in the network. By biasing the sampling distribution, PER effectively changes the expected gradient. 
Our main theoretical contribution shows the expected gradient of a loss function minimized on data sampled with non-uniform probability, is equal to the expected gradient of another, distinct, loss function minimized on data sampled uniformly. This relationship can be used to transform any loss function into a prioritized sampling scheme with a new loss function or vice-versa. We can use this transformation to develop a concrete understanding about the benefits of non-uniform sampling methods such as PER, as well as facilitate the design of novel prioritized methods. Our discoveries can be summarized as follows:

\textbf{Loss function and prioritization should be tied.} The key implication of this result is that the design of prioritized sampling methods should not be considered in isolation from the loss function. We can use this connection to analyze the correctness of methods which use non-uniform sampling by transforming the loss into the uniform-sampling equivalent and considering whether the new loss is in line with the target objective. In particular, we find the PER objective can be unbiased, even without importance sampling, if the loss function is chosen correctly.

\textbf{Variance reduction.} This relationship in expected gradients brings us to a deeper understanding of the benefits of prioritization. We show that the variance of the sampled gradient can be reduced by transforming a uniformly sampled loss function into a carefully chosen prioritization scheme and corresponding loss function. This means that non-uniform sampling is almost always favorable to uniform sampling, at the cost of additional algorithmic and computational complexity.

\textbf{Empirically unnecessary.} While non-uniform sampling is theoretically favorable, in many cases the variance reducing properties will be minimal. 
Perhaps most unexpectedly, we find in a standard benchmark problem, prioritized sampling can be replaced with uniform sampling and a modified loss function, without affecting performance. 
This result suggests some of the benefit of prioritized experience replay comes from the change in expected gradient, rather than the prioritization itself.

We introduce Loss-Adjusted Prioritized (LAP) experience replay and its uniformly sampled loss equivalent, Prioritized Approximation Loss (PAL). LAP simplifies PER by removing unnecessary importance sampling ratios and setting the minimum priority to be $1$, which reduces bias and the likelihood of dead transitions with near-zero sampling probability in a principled manner. On the other hand, our loss function PAL, which resembles a weighted variant of the Huber loss, computes the same expected gradient as LAP, and can be added to any deep reinforcement learning algorithm in only a few lines of code. 
We evaluate both LAP and PAL on the suite of MuJoCo environments \cite{mujoco} and a set of Atari games \cite{bellemare2013arcade}. Across both domains, we find both of our methods outperform the vanilla algorithms they modify. In the MuJoCo domain, we find significant gains over the state-of-the-art in the hardest task, Humanoid. 
All code is open-sourced (\url{https://github.com/sfujim/LAP-PAL}). %

\section{Related Work}

The use of prioritization in reinforcement learning originates from prioritized sweeping for value iteration \cite{moore1993prioritized, andre1998generalized,van2013planning} to increase learning speed, but has also found use in modern applications for importance sampling over trajectories \cite{schlegel2019importance} and learning from demonstrations~\cite{hester2017deep, vevcerik2017leveraging}. Prioritized experience replay \cite{PrioritizedExpReplay} is one of several popular improvements to the DQN algorithm \cite{DQN, DoubleDQN, wang2015dueling, bellemare2017distributional} and has been included in many algorithms  combining multiple improvements~\cite{jaderberg2016reinforcement, hessel2017rainbow, horgan2018distributed, barth-maron2018distributional}.
Variations of PER have been proposed for considering sequences of transitions \cite{gruslys2017reactor,lee2019sample, daley2019reconciling, brittain2019prioritized}, or optimizing the prioritization function \cite{zha2019experience}. Alternate replay buffers have been proposed to favor recent transitions without explicit prioritization \cite{novati2019remember, wang2019towards}. 
The composition and size of the replay buffer has been studied \cite{de2015expreplay, de2016improved, zhang2017expreplay, isele2018selective}, as well as prioritization in simple environments \cite{liu2018effects}.

Non-uniform sampling has been used in the context of training deep networks faster by favoring informative samples~\cite{loshchilov2015online}, or by distributed and non-uniform sampling where data points are re-weighted by importance sampling ratios~\cite{alain2015variance, zhao2015stochastic}. There also exists other importance sampling approaches \cite{needell2014stochastic, katharopoulos2018not} which re-weight updates to reduce variance. In contrast, our method avoids the use of importance sampling by considering changes to the loss function itself and focuses on the context of deep reinforcement learning. 

\section{Preliminaries}
We consider a standard reinforcement learning problem setup, framed by a Markov decision process (MDP), a 5-tuple $(\mathcal{S}, \mathcal{A}, \mathcal{R}, p, \y)$, 
with state space $\mathcal{S}$, action space $\mathcal{A}$, reward function $\mathcal{R}$, dynamics model $p$, and discount factor $\y$.
The behavior of a reinforcement learning agent is defined by its policy $\pi: \mathcal{S} \rightarrow \mathcal{A}$. The performance of a policy $\pi$ can be defined by the value function $Q^\pi$, the expected sum of discounted rewards when following $\pi$ after performing the action $a$ in the state~$s$: $Q^\pi(s,a) = \E_\pi [ \sum_{t=0}^\infty \y^t r_{t+1} | s_0=s, a_0=a ]$.
The value function can be determined from the Bellman equation~\cite{bellman}: $Q^\pi(s, a) = \E_{r,s' \sim p; a' \sim \pi} \lb r + \y Q^\pi(s', a') \rb$. 

In deep reinforcement learning algorithms, such as the Deep Q-Network algorithm (DQN) \cite{watkins1989qlearning,DQN}, the value function is approximated by a neural network $Q_\ta$ with parameters $\ta$. Given transitions $i = (s,a,r,s')$, DQN is trained by minimizing a loss $\Loss(\delta(i))$ on the temporal-difference (TD) error $\delta(i)$ \cite{sutton1988tdlearning}, the difference between the network output $Q(i)=Q_\ta(i)$ and learning target $y(i)$:
\begin{equation} \label{eqn:DQN}
  \delta(i) = Q(i) - y(i), \quad y(i) = r + \gamma Q_{\ta'}(s',a').
\end{equation}
Transitions $i \in \B$ are sampled from an experience replay buffer $\B$ \cite{expreplay1992}, a data set containing previously experienced transitions. 
The target $y(i)$ uses a separate target network $Q_{\ta'}$ with parameters $\ta'$, which are frozen to maintain a fixed target over multiple updates, and then updated to copy the parameters $\ta$ after a set number of learning steps. 
This loss is averaged over a mini-batch of size $M$: $\frac{1}{M} \sum_i \Loss(\delta(i))$. For analysis, the size of $M$ is unimportant as it does not affect the expected loss or gradient. 

Our analysis revolves around the gradient $\g_Q$ of different loss functions with respect to the output of the value network $Q_\ta$, noting $\delta(i)$ is a function of $Q(i)$. In this work, we focus on three loss functions, which we define over the TD error $\delta(i)$ of a transition $i$: the L1 loss $\Loss_\text{L1}(\delta(i)) = |\delta(i)|$ with a gradient $\g_Q \Loss_\text{L1}(\delta(i)) = \text{sign}(\delta(i))$, 
mean squared error (MSE) $\Loss_\text{MSE}(\delta(i)) = 0.5 \delta(i)^2$ with a gradient $\g_Q \Loss_\text{MSE}(\delta(i)) = \delta(i)$,  
and the Huber loss~\cite{huber1964robust}:
\begin{equation}
    \mathcal{L}_\text{Huber}(\delta(i)) = 
    \begin{cases}
    0.5 \delta(i)^2 &\text{if } |\delta(i)| \leq \kappa,\\
    \kappa(|\delta(i)| - 0.5\kappa) &\text{otherwise,}
    \end{cases}
\end{equation}
where generally $\kappa=1$, giving an equivalent gradient to MSE or L1 loss, depending on $|\delta(i)|$.

\textbf{Prioritized Experience Replay.} Prioritized experience replay (PER) \cite{PrioritizedExpReplay} is a non-uniform sampling scheme for replay buffers where transitions are sampled in proportion to their temporal-difference (TD) error. The intuitive argument behind 
PER is that training on the highest error samples will result in the largest performance gain. 

PER makes two changes to the traditional uniformly sampled replay buffer. Firstly, the probability of sampling a transition $i$ is proportionate to the absolute TD error $|\delta(i)|$, set to the power of a hyper-parameter $\al$ to smooth out extremes:
\begin{equation} \label{eqn:PER_probs}
p(i) = \frac{|\delta(i)|^\al + \e}{\sum_j \lp |\delta(j)|^\al + \e \rp},
\end{equation}
where a small constant $\e$ is added to ensure each transition is sampled with non-zero probability. This $\e$ is necessary as often the current TD error is approximated by the TD error when $i$ was last sampled. 
For our theoretical analysis we will treat $\delta(i)$ as the current TD error and assume $\e = 0$.

Secondly, given PER favors transitions with high error, in a stochastic setting it will shift the distribution of $s'$ in the expectation $\E_{s'} \lb Q(s',a') \rb$. This is corrected by weighted importance sampling ratios $w(i)$, which reduce the influence of high priority transitions using a ratio between \autoref{eqn:PER_probs} and the uniform probability $\frac{1}{N}$, where $N$ is the total number of elements contained in the buffer:  
\begin{equation}
\mathcal{L}_\text{PER}(\delta(i)) = w(i) \mathcal{L}(\delta(i)), 
\qquad w(i) = \frac{\hat w(i)}{\max_j \hat w(j)}, 
\qquad \hat w(i) = \lp \frac{1}{N} \cdot \frac{1}{p(i)} \rp^\beta.
\end{equation}
The hyper-parameter $\beta$ is used to smooth out high variance importance sampling weights. The value of $\beta$ is annealed from an initial value $\beta_0$ to $1$, to eliminate any bias from distributional shift.

\section{The Connection between Sampling and Loss Functions} \label{Section:Theory}

In this section, we present our general results which show the expected gradient of a loss function with non-uniform sampling is equivalent to the expected gradient of another loss function with uniform sampling. This relationship provides an approach to analyze methods which use non-uniform sampling, by considering whether the uniformly sampled loss with equivalent expected gradient is reasonable, with regards to the learning objective. All proofs are in the supplementary material.

To build intuition, consider that the expected gradient of a generic loss $\Loss_1$ on the TD error $\delta(i)$ of transitions $i$ sampled by a distribution~$\mathcal{D}_1$, can be determined from another distribution $\mathcal{D}_2$ by using the importance sampling ratio $\frac{p_{\mathcal{D}_1}(i)}{p_{\mathcal{D}_2}(i)}$:
\begin{equation} \label{eqn:example}
    \underbrace{\E_{i \sim \mathcal{D}_1}[\g_Q \Loss_1(\delta(i))]}_{\text{expected gradient of $\Loss_1$ under $\mathcal{D}_1$}} = \E_{i \sim \mathcal{D}_2} \lb \frac{p_{\mathcal{D}_1}(i)}{p_{\mathcal{D}_2}(i)} \g_Q \Loss_1(\delta(i)) \rb.
\end{equation}
Suppose we introduce a second loss $\Loss_2$, where the gradient of $\Loss_2$ is the inner expression of the RHS of \autoref{eqn:example}: $\g_Q \Loss_2(\delta(i)) = \frac{p_{\mathcal{D}_1}(i)}{p_{\mathcal{D}_2}(i)} \g_Q \Loss_1(\delta(i))$. 
Then the expected gradient of $\Loss_1$ under $\mathcal{D}_1$ and $\Loss_2$ under $\mathcal{D}_2$ would be equal $\E_{\mathcal{D}_1}[\g_Q \Loss_1(\delta(i))]=\E_{\mathcal{D}_2}[\g_Q \Loss_2(\delta(i))]$. 
It may first seem unlikely that this relationship would exist in practice. However, defining $\mathcal{D}_1$ to be the uniform distribution~$\mathcal{U}$ over a finite data set $\B$, and $\mathcal{D}_2$ to be a prioritized sampling scheme $p(i) = \frac{|\delta(i)|}{\sum_{j\in \mathcal{B}} |\delta(j)|}$, we have the following relationship between MSE and L1: \begin{equation}
    \underbrace{\E_{\mathcal{U}}[\g_Q \Loss_\text{MSE}(\delta(i))]}_{\text{expected gradient of MSE under $\mathcal{U}$}}
    = \underbrace{ \E_{\mathcal{D}_2} \lb 
    \frac{\sum_j \delta(j)}{N |\delta(i)|}
    \delta(i) \rb}_\text{by \autoref{eqn:example}}
    \propto \E_{\mathcal{D}_2} \underbrace{\lb \text{sign}(\delta(i)) \rb}_{\g_Q \Loss_\text{L1}(\delta(i))}
    = \underbrace{\E_{\mathcal{D}_2}[\g_Q \Loss_\text{L1}(\delta(i))]}_{\text{expected gradient of L1 under $\mathcal{D}_2$}}
\end{equation}
This means the L1 loss, sampled non-uniformly, has the same expected gradient direction as MSE, sampled uniformly.
We emphasize that the distribution $\mathcal{D}$ is very similar to the priority scheme in PER. In the following sections we will generalize this relationship, discuss the benefits to prioritization, and derive a uniformly sampled loss function with the same expected gradient as PER.

We describe our main result in \autoref{Theorem:GradientEquivalence} which formally describes the relationship between a loss $\Loss_1$ on uniformly sampled data, and a loss $\Loss_2$ on data sampled according to some priority scheme $pr$, such that $p(i) = \frac{pr(i)}{\sum_j pr(j)}$. %
As suggested above, this relationship draws a similarity to importance sampling, where the ratio between distributions is absorbed into the loss function. 
\begin{theorem} \label{Theorem:GradientEquivalence}
Given a data set $\B$ of $N$ items, loss functions $\Loss_1$ and $\Loss_2$, and priority scheme $pr$, the expected gradient of $\Loss_1(\delta(i))$, where $i \in \B$ is sampled uniformly, is equal to the expected gradient of $\Loss_2(\delta(i))$, where $i$ is sampled with priority $pr$, if $\g_Q \Loss_1(\delta(i)) = \frac{1}{\lambda} pr(i) \g_Q \Loss_2(\delta(i))$ for all $i$, where~$\lambda = \frac{\sum_j pr(j)}{N}$.
\end{theorem}
\autoref{Theorem:GradientEquivalence} describes the conditions for a uniformly sampled loss and non-uniformly sampled loss to have the same expected gradient. 
This additionally provides a recipe for transforming a given loss function $\Loss_2$ with non-uniform sampling into an equivalent loss function $\Loss_1$ with uniform sampling. %
\begin{corollary}\label{Corollary:Transform}
\autoref{Theorem:GradientEquivalence} is satisfied by any two loss functions $\Loss_1$, where $i \in \B$ is sampled uniformly, and $\Loss_2$, where $i$ is sampled with respect to priority $pr$, if $\Loss_1(\delta(i)) = \frac{1}{\lambda} |pr(i)|_\times \Loss_2(\delta(i))$ for all $i$, where $\lambda = \frac{\sum_j pr(j)}{N}$ and $|\cdot|_\times$ is the stop-gradient operation.
\end{corollary}
As noted by our example in the beginning of the section, often the loss function equivalent of a prioritization method is surprisingly simple.
The converse relationship can also be determined. By carefully choosing how the data is sampled, a loss function $\Loss_1$ can be transformed into (almost) any other loss function $\Loss_2$ in expectation.
\begin{corollary} \label{Corollary:Priority}
\autoref{Theorem:GradientEquivalence} is satisfied by any two loss functions $\Loss_1$, where $i \in \B$ is sampled uniformly, and $\lambda \Loss_2$, where $i$ is sampled with respect to priority $pr$ and $\lambda = \frac{\sum_j pr(j)}{N}$, if $\textnormal{sign}(\g_Q \Loss_1(\delta(i))) = \textnormal{sign}(\g_Q \Loss_2(\delta(i)))$ and $pr(i) = \frac{\g_Q \Loss_1(\delta(i))}{\g_Q \Loss_2(\delta(i))}$ for all $i$.
\end{corollary}
Note that $\text{sign}(\g_Q \Loss_1(\delta(i))) = \text{sign}(\g_Q \Loss_2(\delta(i)))$ is only required as $pr(i)$ must be non-negative, and is trivially satisfied by all loss functions which aim to minimize the distance between the output $Q$ and a given target. In this instance, the non-uniform sampling acts similarly to an importance sampling ratio, re-weighting the gradient of $\Loss_2$ to match the gradient of $\Loss_1$ in expectation. 
\autoref{Corollary:Priority} is perhaps most interesting when $\Loss_2$ is the L1 loss as $\g_Q \Loss_\text{L1}(\delta(i)) = \pm 1$, setting $pr(i) = |\g_Q \Loss_1(\delta(i))|$ allows us to transform the loss $\Loss_1$ into a prioritization scheme. It turns out that transforming a given loss function into its prioritized variant with a L1 loss can be used to reduce the variance of the gradient. 

\begin{observation} \label{Observation:L1Variance}
Given a data set $\B$ of $N$ items and loss function $\Loss_1$, the gradient of the loss function $\lambda \Loss_\textnormal{L1}(\delta(i))$, where $i \in \B$ is sampled with priority $pr(i) = |\g_Q \Loss_1(\delta(i))|$ and $\lambda = \frac{\sum_j pr(j)}{N}$, will have lower (or equal) variance than the gradient of $\Loss_1(\delta(i))$, where $i$ is sampled uniformly.
\end{observation}

In fact, we can generalize this observation one step further, and show that the L1 loss, and corresponding priority scheme, produce the lowest variance of all possible loss functions with the same expected gradient.
\begin{theorem} \label{Theorem:Variance}
Given a data set $\B$ of $N$ items and loss function $\Loss_1$, consider the loss function $\lambda \Loss_2(\delta(i))$, where $i \in \B$ is sampled with priority $pr$ and $\lambda = \frac{\sum_j pr(j)}{N}$, such that \autoref{Theorem:GradientEquivalence} is satisfied. The variance of $\g_Q \lambda \Loss_2(\delta(i))$ 
is minimized when $\Loss_2 = \Loss_\textnormal{L1}$ and $pr(i) = |\g_Q \Loss_1(\delta(i))|$.
\end{theorem}
Intuitively, \autoref{Theorem:Variance} applies by taking equally-sized gradient steps, rather than intermixing small and large steps. 
These results suggest a simple recipe for reducing the variance of any loss function while keeping the expected gradient unchanged, by using the L1 loss and corresponding prioritization.

\section{Corrections to Prioritized Experience Replay}

We now consider prioritized experience replay (PER) \cite{PrioritizedExpReplay} and by following \autoref{Theorem:GradientEquivalence} from the previous section, derive its uniformly sampled equivalent loss function. This relationship allows us to consider corrections and possible simplifications. All proofs are in the supplementary material.

\subsection{An Equivalent Loss Function to Prioritized Experience Replay}

We begin by deriving the uniformly sampled equivalent loss function to PER. Our general finding is that when mean squared error (MSE) is used, including a subset of cases in the Huber loss \cite{huber1964robust}, PER optimizes a loss function on the TD error to a power higher than two. This means PER may favor outliers in its estimate of the expectation in the temporal difference target, rather than learn the mean. Furthermore, we find that the importance sampling ratios used by PER can be absorbed into the loss function themselves, which gives an opportunity to simplify the algorithm.

Firstly, from \autoref{Theorem:GradientEquivalence}, we can derive a general result on a loss function $\frac{1}{\tau}|\delta(i)|^\tau$ when used in conjunction with PER. 
\begin{theorem}
The expected gradient of a loss $\frac{1}{\tau}|\delta(i)|^\tau$, where $\tau > 0$, when used with PER is equal to the expected gradient of the following loss when using a uniformly sampled replay buffer: 
\begin{equation} \label{eqn:PER_tau}
\mathcal{L}_\textnormal{PER}^\tau (\delta(i)) = \frac{\eta N }{\tau + \al - \al\beta} |\delta(i)|^{\tau + \al - \al\beta}, \qquad \eta = \frac{\min_j |\delta(j)|^{\al\beta}}{\sum_j |\delta(j)|^\al}.
\end{equation}
\end{theorem}

Noting that DQN \cite{DQN} traditionally uses the Huber loss, we can now show the form of the uniformly sampled loss function with the same expected gradient as PER, when used with DQN. 

\begin{corollary} \label{thm:PER_loss}
The expected gradient of the Huber loss when used with PER is equal to the expected gradient of the following loss when using a uniformly sampled replay buffer: 
\begin{equation}
    \mathcal{L}_\textnormal{PER}^\textnormal{Huber}(\delta(i)) = \frac{\eta N}{\tau + \al - \al\beta} |\delta(i)|^{\tau + \al - \al\beta}, 
    \qquad \tau =
    \begin{cases}
    2 &\text{if } |\delta(i)| \leq 1,\\
    1 &\text{otherwise,}
    \end{cases}
    \qquad \eta = \frac{\min_j |\delta(j)|^{\al\beta}}{\sum_j |\delta(j)|^\al}.
\end{equation}
\end{corollary}

To understand the significance of \autoref{thm:PER_loss} and what it says about the objective in PER, 
first consider the following two observations on MSE and L1:

\begin{observation} \label{obs:MSE}
(MSE) Let $\B(s,a) \subset \B$ be the subset of transitions containing $(s,a)$ and $\delta(i) = Q(i) - y(i)$. If $\g_Q \E_{i \sim \B(s,a)}[0.5 \delta(i)^2] = 0$ then $Q(s,a) = \textnormal{mean}_{i \in \B(s,a)} y(i)$.
\end{observation}

\begin{observation} \label{obs:L1}
(L1 Loss) Let $\B(s,a) \subset \B$ be the subset of transitions containing $(s,a)$ and $\delta(i) = Q(i) - y(i)$. If $\g_Q \E_{i \sim \B}[|\delta(i)|] = 0$ then $Q(s,a) = \textnormal{median}_{i \in \B(s,a)} y(i)$.
\end{observation}

From \autoref{thm:PER_loss} and the aforementioned observations, we can make several statements:

\textbf{The PER objective is biased if $\tau + \al - \al\beta \neq 2$.} The implication of \autoref{obs:MSE} is that minimizing MSE gives us an estimate of the target of interest, the expected temporal-difference target $y(i) = r + \y \E_{s',a'}[Q(s',a')]$. It follows that optimizing a loss $|\delta(i)|^\tau$ with PER, such that $\tau + \al - \al\beta \neq 2$, produces a biased estimate of the target. On the other hand, we argue that not all bias is equal. From \autoref{obs:L1} we see that minimizing a L1 loss gives the median target, rather than the expectation. Given the effects of function approximation and bootstrapping in deep reinforcement learning, one could argue the median is a reasonable loss function, due to its robustness properties. A likely possibility is that an in-between of MSE and L1 would provide a balance of robustness and ``correctness''. 
Looking at \autoref{eqn:PER_tau}, it turns out this is what PER does when combined with an L1 loss, as the loss will be to a power of $1 + \al - \al\beta \in [1,2]$ for $\al \in (0,1]$ and $\beta \in [0,1]$. However, when combined with MSE, if $\beta < 1$, then $2 + \al - \al\beta > 2$ for $\al \in (0,1]$. 
This means while MSE is normally minimized by the mean, when combined with PER, the loss will be minimized by some expression which may favor outliers. This bias explains the poor performance of PER with standard algorithms in continuous control which rely on MSE \cite{novati2019remember, zha2019experience, fu2019diagnosing, wang2019towards}.

\textbf{Importance sampling can be avoided.} PER uses importance sampling (IS) to re-weight the loss function as an approach to reduce the bias introduced by prioritization. We note that PER with MSE is unbiased if the IS hyper-parameter $\beta = 1$. However, our theoretical results show the prioritization is absorbed into the expected gradient. Consequently, PER can be ``unbiased'' even without using IS ($\beta = 0$) if the expected gradient is still meaningful. As discussed above, this simply means selecting $\al$ and $\tau$ such that $\tau + \al \leq 2$. Furthermore, this allows us to avoid any bias from weighted IS.

\subsection{Loss-Adjusted Prioritized Experience Replay}

We now utilize some of the understanding developed in the previous section for the design of a new prioritized experience replay scheme. We call our variant Loss-Adjusted Prioritized (LAP) experience replay, as our theoretical results argue that the choice of loss function should be closely tied to the choice of prioritization. Mirroring LAP, we also introduce a uniformly sampled loss function, Prioritized Approximation Loss (PAL) with the same expected gradient as LAP. When the variance reduction from prioritization is minimal, PAL can be used as a simple and computationally efficient replacement for LAP, requiring only an adjustment to the loss function used to train the Q-network.

With these ideas in mind, we can now introduce our corrections to the traditional prioritized experience replay, which we combine into a simple prioritization scheme which we call the Loss-Adjusted Prioritized (LAP) experience replay. The simplest variant of LAP would use the L1 loss, as suggested by \autoref{Theorem:Variance}, and sample values with priority $pr(i) = |\delta(i)|^\al$. Following \autoref{Theorem:GradientEquivalence}, we can see the expected gradient of this approach is proportional to $|\delta(i)|^{1+\al}$, when sampled uniformly.

However, in practice, L1 loss may not be preferable as each update takes a constant-sized step, possibly overstepping the target if the learning rate is too large. Instead, we apply the commonly used Huber loss, with $\kappa=1$, which swaps from L1 to MSE when the error falls below a threshold of $1$, scaling the appropriately gradient as $\delta(i)$ approaches $0$. 
When $|\delta(i)| < 1$ and MSE is applied, if we want to avoid the bias introduced from using MSE and prioritization, samples with error below $1$ should be sampled uniformly. 
This can be achieved with a priority scheme $pr(i) = \max(|\delta(i)|^\al,1)$, where samples with otherwise low priority are clipped to be at least $1$. We denote this algorithm LAP and it can be described by non-uniform sampling and the Huber loss: 
\begin{equation}
    p(i) = \frac{\max(|\delta(i)|^\al,1)}{\sum_j \max(|\delta(j)|^\al,1)}, \qquad \Loss_\textnormal{Huber}(\delta(i)) = 
    \begin{cases}
    0.5 \delta(i)^2 &\text{if } |\delta(i)| \leq 1,\\
    |\delta(i)| &\text{otherwise.}
    \end{cases}
\end{equation}
On top of correcting the outlier bias, this clipping reduces the likelihood of dead transitions that occur when $p(i) \approx 0$, eliminating the need for a $\e$ hyper-parameter, which is added to the priority of each transition in PER. LAP maintains the variance reduction properties defined by our theoretical analysis as it uses the L1 loss on all samples with large errors. Following our ideas from the previous section, we can transform LAP into its mirrored loss function with an equivalent expected gradient, which we denote Prioritized Approximation Loss (PAL). PAL can be derived following \autoref{Corollary:Transform}: 
\begin{equation}
    \mathcal{L}_{\text{PAL}}(\delta(i)) = 
    \frac{1}{\lambda} \begin{cases}
    0.5 \delta(i)^2 &\text{if } |\delta(i)| \leq 1,\\
    \frac{|\delta(i)|^{1 + \al}}{1 + \al} &\text{otherwise,}
    \end{cases} \qquad \lambda = \frac{\sum_j \max(|\delta(j)|^\al, 1)}{N}.
\end{equation}

\begin{observation}
LAP and PAL have the same expected gradient. 
\end{observation}

As PAL and LAP share the same expected gradient, we have a mechanism for analysing both methods. Importantly, we note that loss defined by PAL is never to a power greater than two, meaning the outlier bias from PER has been eliminated. %
In some cases, we will find PAL is useful as a loss function on its own, and in domains where the variance reducing property from prioritization is unimportant, we should expect their performance to be similar.

\section{Experiments} \label{section:results}

\begin{figure*}[t]
    \centering
    \includegraphics[width=\linewidth]{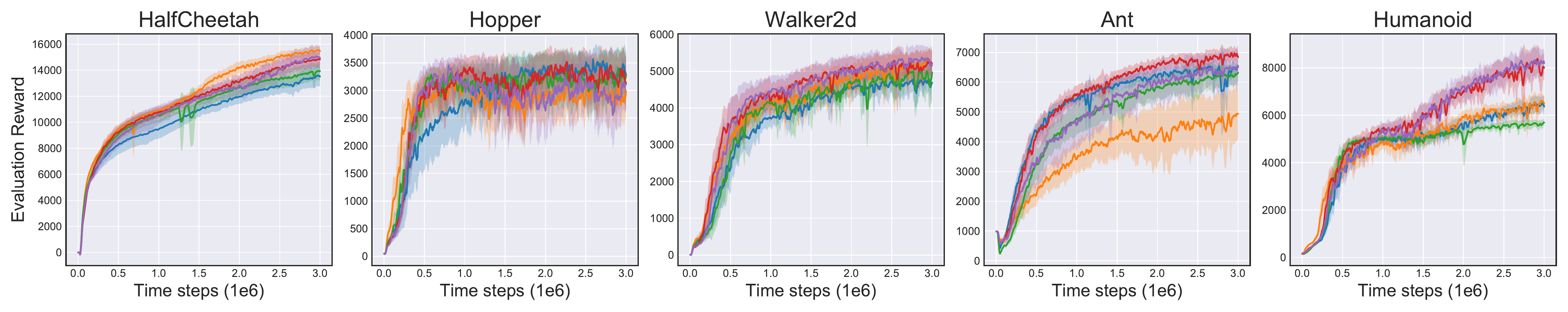}
    \includegraphics[scale=\thelastscalefactor]{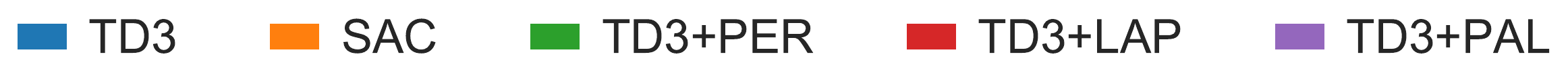}
    \caption{Learning curves for the suite of OpenAI gym continuous control tasks in MuJoCo. Curves are averaged over 10 trials, where the shaded area represents a 95\% confidence interval over the trials.} \label{figure:results} %
    \vskip -0.1in
\end{figure*}

\setlength{\tabcolsep}{3.8pt}
\begin{table}[t]
\centering
\caption{Average performance over the last 10 evaluations and 10 trials. $\pm$ captures a 95\% confidence interval. Scores are bold if the confidence interval intersects with the confidence interval of the highest performance, except for Hopper and Walker2d where all scores satisfy this condition.} \label{table:results}
\small
\begin{tabular}{lccccc}
\toprule
            & TD3                   & SAC               & TD3 + PER         & TD3 + LAP             & TD3 + PAL \\
\midrule
HalfCheetah & 13570.9 $\pm$ 794.2 & \textbf{15511.6 $\pm$ 305.2} & 13927.8 $\pm$ 683.9 & \textbf{14836.5 $\pm$ 532.2} & \textbf{15012.2 $\pm$ 885.4} \\
Hopper      & 3393.2 $\pm$ 381.9 & 2851.6 $\pm$ 417.4 & 3275.5 $\pm$ 451.8 & 3246.9 $\pm$ 463.4 & 3129.1 $\pm$ 473.5 \\
Walker2d    & 4692.4 $\pm$ 423.6 & 5234.4 $\pm$ 346.1 & 4719.1 $\pm$ 492.0 & 5230.5 $\pm$ 368.2 & 5218.7 $\pm$ 422.6 \\
Ant         & 6469.9 $\pm$ 200.3 & 4923.6 $\pm$ 882.3 & 6278.7 $\pm$ 311.3 & \textbf{6912.6 $\pm$ 234.4} & \textbf{6476.2 $\pm$ 640.2} \\
Humanoid    & 6437.5 $\pm$ 349.3 & 6580.9 $\pm$ 296.6 & 5629.3 $\pm$ 174.4 & \textbf{7855.6 $\pm$ 705.9} & \textbf{8265.9 $\pm$ 519.0} \\
\bottomrule
\end{tabular}
\end{table}

\begin{figure}[t]
\centering
\begin{minipage}{.5\textwidth}
\centering
\subfloat{\includegraphics[width=0.5\linewidth]{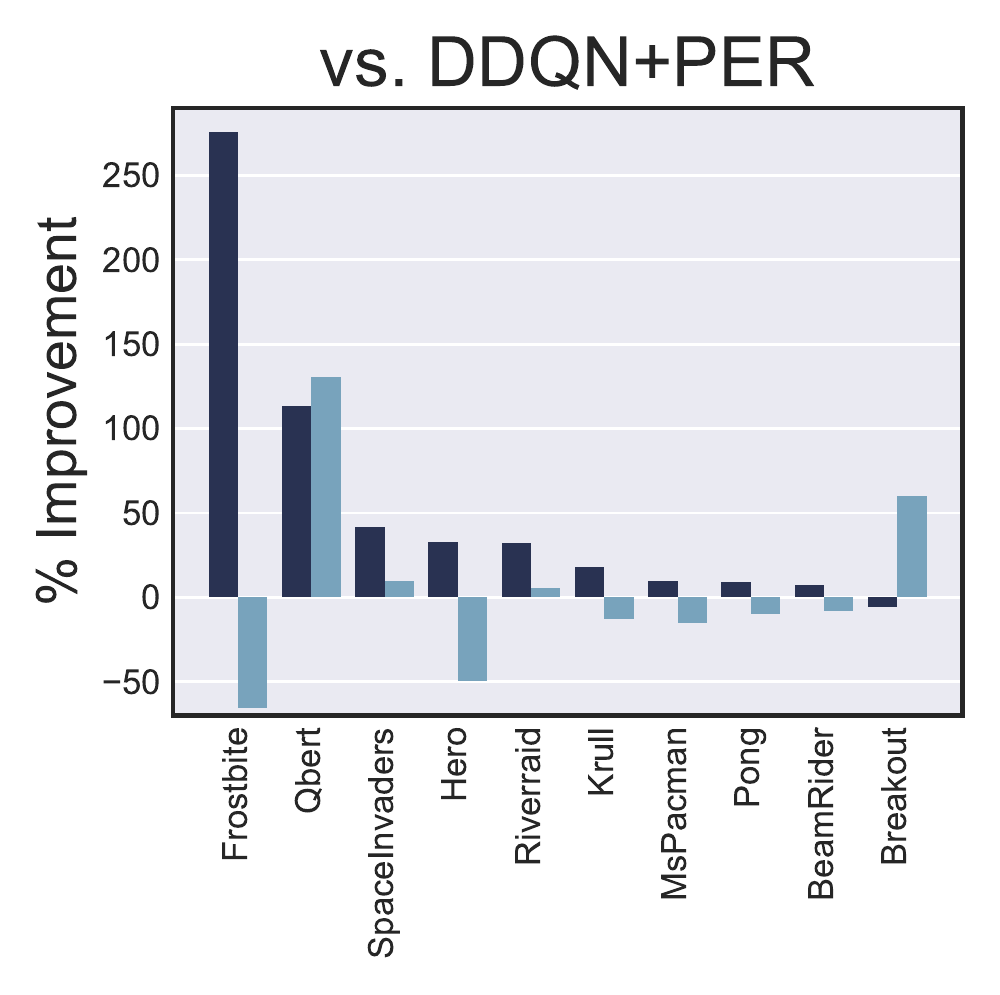}}
\subfloat{\includegraphics[width=0.5\linewidth]{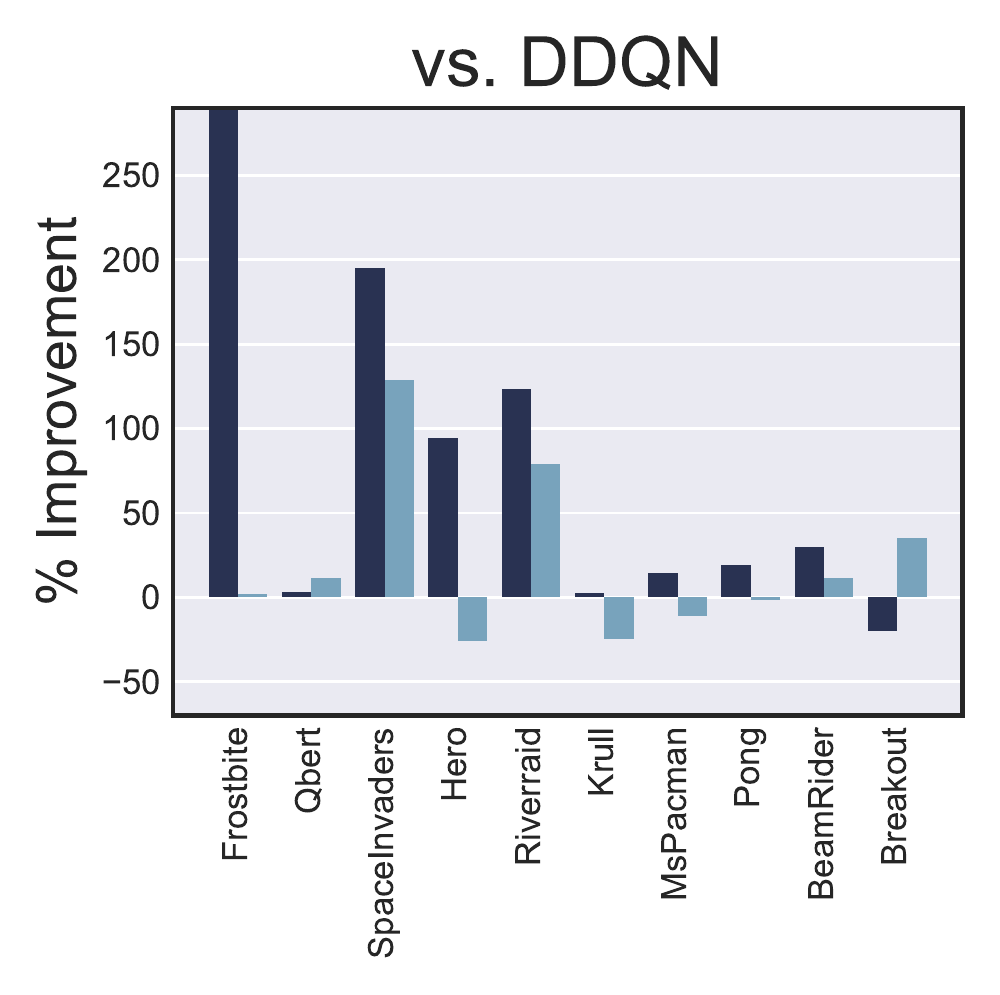}}
\vskip -0.1in
\includegraphics[width=0.5\linewidth]{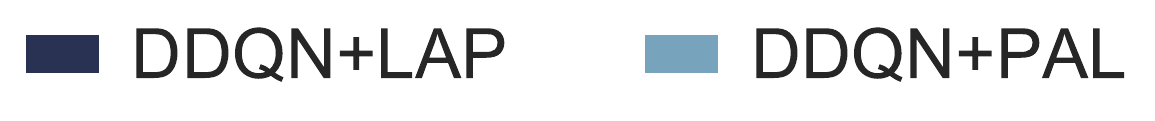}
\captionof{figure}{We display the percentage improvement of final scores achieved by DDQN + PAL and DDQN + LAP when compared to DDQN + PER (left) and DDQN (right). Some extreme values are visually clipped.}
\label{figure:atari}
\end{minipage}\hfill
\begin{minipage}{.47\textwidth}
\centering
\captionof{table}{Mean and median percentage improvement of final scores achieved over DDQN and DDQN + PER across $10$ Atari games.} \label{table:atari}
\small
\setlength{\tabcolsep}{4pt}
\begin{tabular}{lcc}
\toprule
& Mean \% Gain & Median \% Gain \\
\midrule
& \multicolumn{2}{c}{vs. DDQN + PER} \\ 
\cmidrule{2-3}
DDQN & -8.06\% & -13.00\% \\
DDQN + LAP & \textbf{+53.38\%} & \textbf{+24.98\%} \\
DDQN + PAL & +4.50\% & -8.96\% \\
\midrule
& \multicolumn{2}{c}{vs. DDQN} \\ 
\cmidrule{2-3}
DDQN + PER & +37.65\% & +15.24\% \\
DDQN + LAP & \textbf{+148.16\%}  & \textbf{+24.35\%} \\
DDQN + PAL & +20.46\% & +6.79\% \\
\bottomrule
\end{tabular}
\end{minipage}
\end{figure}

We evaluate the benefits of LAP and PAL on the standard suite of MuJoCo \cite{mujoco} continuous control tasks as well as a subset of Atari games, both interfaced through OpenAI gym \cite{OpenAIGym}. 
For continuous control, we combine our methods with a state-of-the-art algorithm TD3 \cite{fujimoto2018addressing}, which we benchmark against, as well as SAC \cite{haarnoja2018soft}, which have both been shown to outperform other standard algorithms on the MuJoCo benchmark. 
For Atari, we apply LAP and PAL to Double DQN (DDQN) \cite{DoubleDQN}, and benchmark the performance against PER + DDQN. 
A complete list of hyper-parameters and experimental details are provided in the supplementary material. MuJoCo results are presented in \autoref{figure:results} and \autoref{table:results} and Atari results are presented in \autoref{figure:atari} and Table \ref{table:atari}.

We find that the addition of either LAP or PAL matches or outperforms the vanilla version of TD3 in all tasks. In the challenging Humanoid task, we find that both LAP and PAL offer a large improvement over the previous state-of-the-art. Interestingly enough, we find no meaningful difference in the performance of LAP and PAL across all tasks. This means that prioritization has little benefit
and the improvement comes from the change in expected gradient from using our methods over MSE. Consequently, for MuJoCo environments, non-uniform sampling can be replaced by adjusting the loss function instead. Additionally, we confirm previous results which found PER provides no benefit when added to TD3 \cite{fu2019diagnosing}. Given prioritization appears to have little impact in this domain, this result is consistent with our theoretical analysis which shows using that MSE with PER introduces bias. 

In the Atari domain, we find LAP improves over PER in $9$ out of $10$ environments, with a mean performance gain of $53\%$. On the other hand, while PAL offers non-trivial gains over vanilla DDQN, it under-performs PER on $6$ of the $10$ environments tested. This suggests that prioritization plays a more significant role in the Atari domain, but some of the improvement can still be attributed to the change in expected gradient. As the Atari domain includes games which require longer horizons or sparse rewards, the benefit of prioritization is not surprising.

When $\al=0$, PAL equals a Huber loss with a scaling factor $\frac{1}{\lambda}$. We perform an ablation study to disentangle the importance of the components to PAL. We compare the performance of PAL and the Huber loss with and without $\frac{1}{\lambda}$. The results are presented in the supplementary material. We find the complete loss of PAL achieves the highest performance, and the change in expected gradient from using $\al \neq 0$ to be the largest contributing factor.

\section{Discussion}

\textbf{Performance.} The performance of PAL is of particular interest due to the simplicity of the method, requiring only a small change to the loss function. In the MuJoCo domain, TD3 + PAL matches the performance of our prioritization scheme while outperforming vanilla TD3 and SAC. Our ablation study shows the performance gain largely comes from the change in expected gradient. We believe the benefit of PAL over MSE is the additional robustness, and the benefit over the Huber loss is a better approximation of the mean.
There is also a small gain in performance due to the re-scaling used to match the exact expected gradient of LAP. When examining PAL in isolation, it is somewhat unclear why this re-scaling should improve performance. One hypothesis is that as $\frac{1}{\lambda}$ decreases the average error increases, this can be thought of an approach to balance out large gradient steps. On the other hand, the performance gain from LAP is more intuitive to understand, as we correct underlying issues with PER by understanding its relationship with the loss function.

\textbf{Alternate Priority schemes.} Our theoretical results define a relationship between the expected gradient of a non-uniformly sampled loss function and a uniformly sampled loss function. This allows us define the optimal variance-reducing prioritization scheme and corresponding loss function. 
However, there is still more research to do in non-uniform sampling, as often the a uniform sample of the replay buffer is not representative of the data of interest \cite{fujimoto2019off}, and alternate weightings may be preferred such as the stationary distribution~\cite{liu2018breaking, liu2019off, nachum2019dualdice} or importance sampling ratios \cite{schlegel2019importance}. We hope our results provide a valuable tool for analyzing and understanding alternate prioritization schemes. 

\textbf{Reproducibility and algorithmic credit assignment.} Our work emphasizes the susceptible nature of deep reinforcement learning algorithms to small changes~\cite{ilyas2018deep} and the reproducibility crisis~\cite{hendersonRL2017}, as we are able to show significant improvements to the performance of a well-known algorithm with only minor changes to the loss function. This suggests that papers which use intensive hyper-parameter optimization or introduce algorithmic changes without ablation studies may be improving over the original algorithm due to unintended consequences, rather than the proposed method.

\section{Conclusion}

In this paper, we aim to build the theoretical foundations for a commonly used deep reinforcement learning technique known as prioritized experience replay (PER) \cite{PrioritizedExpReplay}. To do so, 
we first show an interesting connection between non-uniform sampling and loss functions. Namely, any loss function can be approximated, in the sense of having the same expected gradient, by a new loss function with some non-uniform sampling scheme, and vice-versa. We use this relationship to show that the prioritized, non-uniformly sampled variant of the loss function has lower variance than the uniformly sampled equivalent. This result suggests that by carefully considering the loss function, using prioritization should outperform the standard setup of uniform sampling.

However, without considering the loss function, prioritization changes the expected gradient. This allows us to 
develop a concrete understanding of why PER has been shown to perform poorly when combined with continuous control methods in the past~\cite{fu2019diagnosing, wang2019towards}, due to a bias towards outliers when used with MSE. 
We introduce a corrected version of PER which considers the loss function, known as Loss-Adjusted Prioritized (LAP) experience replay and its mirrored uniformly sampled loss function equivalent, Prioritized Approximation Loss (PAL). 
We test both LAP and PAL on standard deep reinforcement learning benchmarks in MuJoCo and Atari, and show their addition improves upon the vanilla algorithm across both domains.

\section*{Broader Impact}

Our research focuses on developing the theoretical foundations for a commonly used technique in deep reinforcement learning, prioritized experience replay. Our insights could be used to improve reinforcement learning systems over a wide range of applications such as clinical trial design \cite{zhao2009reinforcement}, educational games \cite{mandel2014offline} and recommender systems \cite{swaminathan2017off,gauci2018horizon}. Non-uniform sampling may also have benefits for scaling offline reinforcement learning, in particular, when learning from large data sets where sampling only relevant or important data is critical.
We expect our impact to be more significant for the reinforcement learning community itself. In the supplementary material we demonstrate our publicly released implementation of PER runs significantly faster (5-17$\times$) than previous implementations published by corporate research groups \cite{baselines, castro2018dopamine}. This improves the accessibility of non-uniform sampling strategy and state-of-the-art deep reinforcement learning research for groups with resource limitations.

\begin{ack}
Scott Fujimoto is supported by a NSERC scholarship as well as the Borealis AI Global Fellowship Award. This research was enabled in part by support provided by Calcul Qu\'ebec and Compute Canada. We would like to thank Edward Smith for helpful discussions and feedback.
\end{ack}

\bibliography{example_paper}

\begin{thebibliography}{61}
\providecommand{\natexlab}[1]{#1}
\providecommand{\url}[1]{\texttt{#1}}
\expandafter\ifx\csname urlstyle\endcsname\relax
  \providecommand{\doi}[1]{doi: #1}\else
  \providecommand{\doi}{doi: \begingroup \urlstyle{rm}\Url}\fi

\bibitem[Schaul et~al.(2016)Schaul, Quan, Antonoglou, and
  Silver]{PrioritizedExpReplay}
Tom Schaul, John Quan, Ioannis Antonoglou, and David Silver.
\newblock Prioritized experience replay.
\newblock In \emph{International Conference on Learning Representations},
  Puerto Rico, 2016.

\bibitem[Hessel et~al.(2017)Hessel, Modayil, Van~Hasselt, Schaul, Ostrovski,
  Dabney, Horgan, Piot, Azar, and Silver]{hessel2017rainbow}
Matteo Hessel, Joseph Modayil, Hado Van~Hasselt, Tom Schaul, Georg Ostrovski,
  Will Dabney, Dan Horgan, Bilal Piot, Mohammad Azar, and David Silver.
\newblock Rainbow: Combining improvements in deep reinforcement learning.
\newblock \emph{arXiv preprint arXiv:1710.02298}, 2017.

\bibitem[Todorov et~al.(2012)Todorov, Erez, and Tassa]{mujoco}
Emanuel Todorov, Tom Erez, and Yuval Tassa.
\newblock Mujoco: A physics engine for model-based control.
\newblock In \emph{IEEE/RSJ International Conference on Intelligent Robots and
  Systems (IROS)}, pages 5026--5033. IEEE, 2012.

\bibitem[Bellemare et~al.(2013)Bellemare, Naddaf, Veness, and
  Bowling]{bellemare2013arcade}
Marc~G Bellemare, Yavar Naddaf, Joel Veness, and Michael Bowling.
\newblock The arcade learning environment: An evaluation platform for general
  agents.
\newblock \emph{Journal of Artificial Intelligence Research}, 47:\penalty0
  253--279, 2013.

\bibitem[Moore and Atkeson(1993)]{moore1993prioritized}
Andrew~W Moore and Christopher~G Atkeson.
\newblock Prioritized sweeping: Reinforcement learning with less data and less
  time.
\newblock \emph{Machine learning}, 13\penalty0 (1):\penalty0 103--130, 1993.

\bibitem[Andre et~al.(1998)Andre, Friedman, and Parr]{andre1998generalized}
David Andre, Nir Friedman, and Ronald Parr.
\newblock Generalized prioritized sweeping.
\newblock In \emph{Advances in Neural Information Processing Systems}, pages
  1001--1007, 1998.

\bibitem[Van~Seijen and Sutton(2013)]{van2013planning}
Harm Van~Seijen and Richard~S Sutton.
\newblock Planning by prioritized sweeping with small backups.
\newblock \emph{arXiv preprint arXiv:1301.2343}, 2013.

\bibitem[Schlegel et~al.(2019)Schlegel, Chung, Graves, Qian, and
  White]{schlegel2019importance}
Matthew Schlegel, Wesley Chung, Daniel Graves, Jian Qian, and Martha White.
\newblock Importance resampling for off-policy prediction.
\newblock In \emph{Advances in Neural Information Processing Systems}, pages
  1797--1807, 2019.

\bibitem[Hester et~al.(2017)Hester, Vecerik, Pietquin, Lanctot, Schaul, Piot,
  Horgan, Quan, Sendonaris, Dulac-Arnold, et~al.]{hester2017deep}
Todd Hester, Matej Vecerik, Olivier Pietquin, Marc Lanctot, Tom Schaul, Bilal
  Piot, Dan Horgan, John Quan, Andrew Sendonaris, Gabriel Dulac-Arnold, et~al.
\newblock Deep q-learning from demonstrations.
\newblock \emph{arXiv preprint arXiv:1704.03732}, 2017.

\bibitem[Ve{\v{c}}er{\'\i}k et~al.(2017)Ve{\v{c}}er{\'\i}k, Hester, Scholz,
  Wang, Pietquin, Piot, Heess, Roth{\"o}rl, Lampe, and
  Riedmiller]{vevcerik2017leveraging}
Matej Ve{\v{c}}er{\'\i}k, Todd Hester, Jonathan Scholz, Fumin Wang, Olivier
  Pietquin, Bilal Piot, Nicolas Heess, Thomas Roth{\"o}rl, Thomas Lampe, and
  Martin Riedmiller.
\newblock Leveraging demonstrations for deep reinforcement learning on robotics
  problems with sparse rewards.
\newblock \emph{arXiv preprint arXiv:1707.08817}, 2017.

\bibitem[Mnih et~al.(2015)Mnih, Kavukcuoglu, Silver, Rusu, Veness, Bellemare,
  Graves, Riedmiller, Fidjeland, Ostrovski, et~al.]{DQN}
Volodymyr Mnih, Koray Kavukcuoglu, David Silver, Andrei~A Rusu, Joel Veness,
  Marc~G Bellemare, Alex Graves, Martin Riedmiller, Andreas~K Fidjeland, Georg
  Ostrovski, et~al.
\newblock Human-level control through deep reinforcement learning.
\newblock \emph{Nature}, 518\penalty0 (7540):\penalty0 529--533, 2015.

\bibitem[Van~Hasselt et~al.(2016)Van~Hasselt, Guez, and Silver]{DoubleDQN}
Hado Van~Hasselt, Arthur Guez, and David Silver.
\newblock Deep reinforcement learning with double q-learning.
\newblock In \emph{Thirtieth AAAI conference on artificial intelligence}, 2016.

\bibitem[Wang et~al.(2016)Wang, Schaul, Hessel, Hasselt, Lanctot, and
  Freitas]{wang2015dueling}
Ziyu Wang, Tom Schaul, Matteo Hessel, Hado Hasselt, Marc Lanctot, and Nando
  Freitas.
\newblock Dueling network architectures for deep reinforcement learning.
\newblock In \emph{International Conference on Machine Learning}, pages
  1995--2003, 2016.

\bibitem[Bellemare et~al.(2017)Bellemare, Dabney, and
  Munos]{bellemare2017distributional}
Marc~G Bellemare, Will Dabney, and R{\'e}mi Munos.
\newblock A distributional perspective on reinforcement learning.
\newblock In \emph{International Conference on Machine Learning}, pages
  449--458, 2017.

\bibitem[Jaderberg et~al.(2016)Jaderberg, Mnih, Czarnecki, Schaul, Leibo,
  Silver, and Kavukcuoglu]{jaderberg2016reinforcement}
Max Jaderberg, Volodymyr Mnih, Wojciech~Marian Czarnecki, Tom Schaul, Joel~Z
  Leibo, David Silver, and Koray Kavukcuoglu.
\newblock Reinforcement learning with unsupervised auxiliary tasks.
\newblock \emph{arXiv preprint arXiv:1611.05397}, 2016.

\bibitem[Horgan et~al.(2018)Horgan, Quan, Budden, Barth-Maron, Hessel, van
  Hasselt, and Silver]{horgan2018distributed}
Dan Horgan, John Quan, David Budden, Gabriel Barth-Maron, Matteo Hessel, Hado
  van Hasselt, and David Silver.
\newblock Distributed prioritized experience replay.
\newblock \emph{International Conference on Learning Representations}, 2018.

\bibitem[Barth-Maron et~al.(2018)Barth-Maron, Hoffman, Budden, Dabney, Horgan,
  TB, Muldal, Heess, and Lillicrap]{barth-maron2018distributional}
Gabriel Barth-Maron, Matthew~W Hoffman, David Budden, Will Dabney, Dan Horgan,
  Dhruva TB, Alistair Muldal, Nicolas Heess, and Timothy Lillicrap.
\newblock Distributional policy gradients.
\newblock \emph{International Conference on Learning Representations}, 2018.

\bibitem[Gruslys et~al.(2017)Gruslys, Dabney, Azar, Piot, Bellemare, and
  Munos]{gruslys2017reactor}
Audrunas Gruslys, Will Dabney, Mohammad~Gheshlaghi Azar, Bilal Piot, Marc
  Bellemare, and Remi Munos.
\newblock The reactor: A fast and sample-efficient actor-critic agent for
  reinforcement learning.
\newblock \emph{arXiv preprint arXiv:1704.04651}, 2017.

\bibitem[Lee et~al.(2019)Lee, Sungik, and Chung]{lee2019sample}
Su~Young Lee, Choi Sungik, and Sae-Young Chung.
\newblock Sample-efficient deep reinforcement learning via episodic backward
  update.
\newblock In \emph{Advances in Neural Information Processing Systems}, pages
  2110--2119, 2019.

\bibitem[Daley and Amato(2019)]{daley2019reconciling}
Brett Daley and Christopher Amato.
\newblock Reconciling $\lambda$-returns with experience replay.
\newblock In \emph{Advances in Neural Information Processing Systems 32}, pages
  1131--1140, 2019.

\bibitem[Brittain et~al.(2019)Brittain, Bertram, Yang, and
  Wei]{brittain2019prioritized}
Marc Brittain, Josh Bertram, Xuxi Yang, and Peng Wei.
\newblock Prioritized sequence experience replay.
\newblock \emph{arXiv preprint arXiv:1905.12726}, 2019.

\bibitem[Zha et~al.(2019)Zha, Lai, Zhou, and Hu]{zha2019experience}
Daochen Zha, Kwei-Herng Lai, Kaixiong Zhou, and Xia Hu.
\newblock Experience replay optimization.
\newblock In \emph{Proceedings of the 28th International Joint Conference on
  Artificial Intelligence}, pages 4243--4249. AAAI Press, 2019.

\bibitem[Novati and Koumoutsakos(2019)]{novati2019remember}
Guido Novati and Petros Koumoutsakos.
\newblock Remember and forget for experience replay.
\newblock In \emph{International Conference on Machine Learning}, pages
  4851--4860, 2019.

\bibitem[Wang et~al.(2019)Wang, Wu, Vuong, and Ross]{wang2019towards}
Che Wang, Yanqiu Wu, Quan Vuong, and Keith Ross.
\newblock Towards simplicity in deep reinforcement learning: Streamlined
  off-policy learning.
\newblock \emph{arXiv preprint arXiv:1910.02208}, 2019.

\bibitem[de~Bruin et~al.(2015)de~Bruin, Kober, Tuyls, and
  Babu{\v{s}}ka]{de2015expreplay}
Tim de~Bruin, Jens Kober, Karl Tuyls, and Robert Babu{\v{s}}ka.
\newblock The importance of experience replay database composition in deep
  reinforcement learning.
\newblock In \emph{Deep Reinforcement Learning Workshop, NIPS}, 2015.

\bibitem[de~Bruin et~al.(2016)de~Bruin, Kober, Tuyls, and
  Babu{\v{s}}ka]{de2016improved}
Tim de~Bruin, Jens Kober, Karl Tuyls, and Robert Babu{\v{s}}ka.
\newblock Improved deep reinforcement learning for robotics through
  distribution-based experience retention.
\newblock In \emph{IEEE/RSJ International Conference on Intelligent Robots and
  Systems (IROS)}, pages 3947--3952. IEEE, 2016.

\bibitem[Zhang and Sutton(2017)]{zhang2017expreplay}
Shangtong Zhang and Richard~S Sutton.
\newblock A deeper look at experience replay.
\newblock \emph{arXiv preprint arXiv:1712.01275}, 2017.

\bibitem[Isele and Cosgun(2018)]{isele2018selective}
David Isele and Akansel Cosgun.
\newblock Selective experience replay for lifelong learning.
\newblock \emph{arXiv preprint arXiv:1802.10269}, 2018.

\bibitem[Liu and Zou(2018)]{liu2018effects}
Ruishan Liu and James Zou.
\newblock The effects of memory replay in reinforcement learning.
\newblock In \emph{2018 56th Annual Allerton Conference on Communication,
  Control, and Computing (Allerton)}, pages 478--485. IEEE, 2018.

\bibitem[Loshchilov and Hutter(2015)]{loshchilov2015online}
Ilya Loshchilov and Frank Hutter.
\newblock Online batch selection for faster training of neural networks.
\newblock \emph{arXiv preprint arXiv:1511.06343}, 2015.

\bibitem[Alain et~al.(2015)Alain, Lamb, Sankar, Courville, and
  Bengio]{alain2015variance}
Guillaume Alain, Alex Lamb, Chinnadhurai Sankar, Aaron Courville, and Yoshua
  Bengio.
\newblock Variance reduction in sgd by distributed importance sampling.
\newblock \emph{arXiv preprint arXiv:1511.06481}, 2015.

\bibitem[Zhao and Zhang(2015)]{zhao2015stochastic}
Peilin Zhao and Tong Zhang.
\newblock Stochastic optimization with importance sampling for regularized loss
  minimization.
\newblock In \emph{international conference on machine learning}, pages 1--9,
  2015.

\bibitem[Needell et~al.(2014)Needell, Ward, and Srebro]{needell2014stochastic}
Deanna Needell, Rachel Ward, and Nati Srebro.
\newblock Stochastic gradient descent, weighted sampling, and the randomized
  kaczmarz algorithm.
\newblock In \emph{Advances in neural information processing systems}, pages
  1017--1025, 2014.

\bibitem[Katharopoulos and Fleuret(2018)]{katharopoulos2018not}
Angelos Katharopoulos and Francois Fleuret.
\newblock Not all samples are created equal: Deep learning with importance
  sampling.
\newblock In \emph{International Conference on Machine Learning}, pages
  2525--2534, 2018.

\bibitem[Bellman(1957)]{bellman}
Richard Bellman.
\newblock \emph{Dynamic Programming}.
\newblock Princeton University Press, 1957.

\bibitem[Watkins(1989)]{watkins1989qlearning}
Christopher John Cornish~Hellaby Watkins.
\newblock \emph{Learning from delayed rewards}.
\newblock PhD thesis, King's College, Cambridge, 1989.

\bibitem[Sutton(1988)]{sutton1988tdlearning}
Richard~S Sutton.
\newblock Learning to predict by the methods of temporal differences.
\newblock \emph{Machine learning}, 3\penalty0 (1):\penalty0 9--44, 1988.

\bibitem[Lin(1992)]{expreplay1992}
Long-Ji Lin.
\newblock Self-improving reactive agents based on reinforcement learning,
  planning and teaching.
\newblock \emph{Machine learning}, 8\penalty0 (3-4):\penalty0 293--321, 1992.

\bibitem[Huber et~al.(1964)]{huber1964robust}
Peter~J Huber et~al.
\newblock Robust estimation of a location parameter.
\newblock \emph{The annals of mathematical statistics}, 35\penalty0
  (1):\penalty0 73--101, 1964.

\bibitem[Fu et~al.(2019)Fu, Kumar, Soh, and Levine]{fu2019diagnosing}
Justin Fu, Aviral Kumar, Matthew Soh, and Sergey Levine.
\newblock Diagnosing bottlenecks in deep q-learning algorithms.
\newblock In \emph{International Conference on Machine Learning}, pages
  2021--2030, 2019.

\bibitem[Brockman et~al.(2016)Brockman, Cheung, Pettersson, Schneider,
  Schulman, Tang, and Zaremba]{OpenAIGym}
Greg Brockman, Vicki Cheung, Ludwig Pettersson, Jonas Schneider, John Schulman,
  Jie Tang, and Wojciech Zaremba.
\newblock Openai gym, 2016.

\bibitem[Fujimoto et~al.(2018)Fujimoto, van Hoof, and
  Meger]{fujimoto2018addressing}
Scott Fujimoto, Herke van Hoof, and David Meger.
\newblock Addressing function approximation error in actor-critic methods.
\newblock In \emph{International Conference on Machine Learning}, volume~80,
  pages 1587--1596. PMLR, 2018.

\bibitem[Haarnoja et~al.(2018{\natexlab{a}})Haarnoja, Zhou, Abbeel, and
  Levine]{haarnoja2018soft}
Tuomas Haarnoja, Aurick Zhou, Pieter Abbeel, and Sergey Levine.
\newblock Soft actor-critic: Off-policy maximum entropy deep reinforcement
  learning with a stochastic actor.
\newblock In \emph{International Conference on Machine Learning}, volume~80,
  pages 1861--1870. PMLR, 2018{\natexlab{a}}.

\bibitem[Fujimoto et~al.(2019{\natexlab{a}})Fujimoto, Meger, and
  Precup]{fujimoto2019off}
Scott Fujimoto, David Meger, and Doina Precup.
\newblock Off-policy deep reinforcement learning without exploration.
\newblock In \emph{International Conference on Machine Learning}, pages
  2052--2062, 2019{\natexlab{a}}.

\bibitem[Liu et~al.(2018)Liu, Li, Tang, and Zhou]{liu2018breaking}
Qiang Liu, Lihong Li, Ziyang Tang, and Dengyong Zhou.
\newblock Breaking the curse of horizon: Infinite-horizon off-policy
  estimation.
\newblock In \emph{Advances in Neural Information Processing Systems}, pages
  5356--5366, 2018.

\bibitem[Liu et~al.(2019)Liu, Swaminathan, Agarwal, and Brunskill]{liu2019off}
Yao Liu, Adith Swaminathan, Alekh Agarwal, and Emma Brunskill.
\newblock Off-policy policy gradient with state distribution correction.
\newblock \emph{arXiv preprint arXiv:1904.08473}, 2019.

\bibitem[Nachum et~al.(2019)Nachum, Chow, Dai, and Li]{nachum2019dualdice}
Ofir Nachum, Yinlam Chow, Bo~Dai, and Lihong Li.
\newblock Dualdice: Behavior-agnostic estimation of discounted stationary
  distribution corrections.
\newblock In \emph{Advances in Neural Information Processing Systems}, pages
  2315--2325, 2019.

\bibitem[Ilyas et~al.(2018)Ilyas, Engstrom, Santurkar, Tsipras, Janoos,
  Rudolph, and Madry]{ilyas2018deep}
Andrew Ilyas, Logan Engstrom, Shibani Santurkar, Dimitris Tsipras, Firdaus
  Janoos, Larry Rudolph, and Aleksander Madry.
\newblock Are deep policy gradient algorithms truly policy gradient algorithms?
\newblock \emph{arXiv preprint arXiv:1811.02553}, 2018.

\bibitem[Henderson et~al.(2018)Henderson, Islam, Bachman, Pineau, Precup, and
  Meger]{hendersonRL2017}
Peter Henderson, Riashat Islam, Philip Bachman, Joelle Pineau, Doina Precup,
  and David Meger.
\newblock Deep reinforcement learning that matters.
\newblock In \emph{Thirty-Second AAAI Conference on Artificial Intelligence},
  2018.

\bibitem[Zhao et~al.(2009)Zhao, Kosorok, and Zeng]{zhao2009reinforcement}
Yufan Zhao, Michael~R Kosorok, and Donglin Zeng.
\newblock Reinforcement learning design for cancer clinical trials.
\newblock \emph{Statistics in medicine}, 28\penalty0 (26):\penalty0 3294, 2009.

\bibitem[Mandel et~al.(2014)Mandel, Liu, Levine, Brunskill, and
  Popovic]{mandel2014offline}
Travis Mandel, Yun-En Liu, Sergey Levine, Emma Brunskill, and Zoran Popovic.
\newblock Offline policy evaluation across representations with applications to
  educational games.
\newblock In \emph{International Conference on Autonomous Agents and Multiagent
  Systems}, 2014.

\bibitem[Swaminathan et~al.(2017)Swaminathan, Krishnamurthy, Agarwal, Dudik,
  Langford, Jose, and Zitouni]{swaminathan2017off}
Adith Swaminathan, Akshay Krishnamurthy, Alekh Agarwal, Miro Dudik, John
  Langford, Damien Jose, and Imed Zitouni.
\newblock Off-policy evaluation for slate recommendation.
\newblock In \emph{Advances in Neural Information Processing Systems}, pages
  3632--3642, 2017.

\bibitem[Gauci et~al.(2018)Gauci, Conti, Liang, Virochsiri, He, Kaden,
  Narayanan, Ye, Chen, and Fujimoto]{gauci2018horizon}
Jason Gauci, Edoardo Conti, Yitao Liang, Kittipat Virochsiri, Yuchen He,
  Zachary Kaden, Vivek Narayanan, Xiaohui Ye, Zhengxing Chen, and Scott
  Fujimoto.
\newblock Horizon: Facebook's open source applied reinforcement learning
  platform.
\newblock \emph{arXiv preprint arXiv:1811.00260}, 2018.

\bibitem[Dhariwal et~al.(2017)Dhariwal, Hesse, Plappert, Radford, Schulman,
  Sidor, and Wu]{baselines}
Prafulla Dhariwal, Christopher Hesse, Matthias Plappert, Alec Radford, John
  Schulman, Szymon Sidor, and Yuhuai Wu.
\newblock Openai baselines.
\newblock \url{https://github.com/openai/baselines}, 2017.

\bibitem[Castro et~al.(2018)Castro, Moitra, Gelada, Kumar, and
  Bellemare]{castro2018dopamine}
Pablo~Samuel Castro, Subhodeep Moitra, Carles Gelada, Saurabh Kumar, and Marc~G
  Bellemare.
\newblock Dopamine: A research framework for deep reinforcement learning.
\newblock \emph{arXiv preprint arXiv:1812.06110}, 2018.

\bibitem[Paszke et~al.(2019)Paszke, Gross, Massa, Lerer, Bradbury, Chanan,
  Killeen, Lin, Gimelshein, Antiga, et~al.]{paszke2019pytorch}
Adam Paszke, Sam Gross, Francisco Massa, Adam Lerer, James Bradbury, Gregory
  Chanan, Trevor Killeen, Zeming Lin, Natalia Gimelshein, Luca Antiga, et~al.
\newblock Pytorch: An imperative style, high-performance deep learning library.
\newblock In \emph{Advances in Neural Information Processing Systems}, pages
  8024--8035, 2019.

\bibitem[Kingma and Ba(2014)]{adam}
Diederik Kingma and Jimmy Ba.
\newblock Adam: A method for stochastic optimization.
\newblock \emph{arXiv preprint arXiv:1412.6980}, 2014.

\bibitem[Lillicrap et~al.(2015)Lillicrap, Hunt, Pritzel, Heess, Erez, Tassa,
  Silver, and Wierstra]{DDPG}
Timothy~P Lillicrap, Jonathan~J Hunt, Alexander Pritzel, Nicolas Heess, Tom
  Erez, Yuval Tassa, David Silver, and Daan Wierstra.
\newblock Continuous control with deep reinforcement learning.
\newblock \emph{arXiv preprint arXiv:1509.02971}, 2015.

\bibitem[Haarnoja et~al.(2018{\natexlab{b}})Haarnoja, Zhou, Hartikainen,
  Tucker, Ha, Tan, Kumar, Zhu, Gupta, Abbeel, and
  Levine]{haarnoja2018applications}
Tuomas Haarnoja, Aurick Zhou, Kristian Hartikainen, George Tucker, Sehoon Ha,
  Jie Tan, Vikash Kumar, Henry Zhu, Abhishek Gupta, Pieter Abbeel, and Sergey
  Levine.
\newblock Soft actor-critic algorithms and applications.
\newblock \emph{arXiv preprint arXiv:1812.05905}, 2018{\natexlab{b}}.

\bibitem[Machado et~al.(2018)Machado, Bellemare, Talvitie, Veness, Hausknecht,
  and Bowling]{machado2018revisiting}
Marlos~C Machado, Marc~G Bellemare, Erik Talvitie, Joel Veness, Matthew
  Hausknecht, and Michael Bowling.
\newblock Revisiting the arcade learning environment: Evaluation protocols and
  open problems for general agents.
\newblock \emph{Journal of Artificial Intelligence Research}, 61:\penalty0
  523--562, 2018.

\bibitem[Fujimoto et~al.(2019{\natexlab{b}})Fujimoto, Conti, Ghavamzadeh, and
  Pineau]{fujimoto2019benchmarking}
Scott Fujimoto, Edoardo Conti, Mohammad Ghavamzadeh, and Joelle Pineau.
\newblock Benchmarking batch deep reinforcement learning algorithms.
\newblock \emph{arXiv preprint arXiv:1910.01708}, 2019{\natexlab{b}}.

\end{thebibliography}
\bibliographystyle{unsrtnat}

\clearpage

\appendix

\section{Detailed Proofs}

\setcounter{theorem}{0}
\setcounter{corollary}{0}
\setcounter{observation}{0}
\setcounter{lemma}{0}

\subsection{Theorem 1}

\begin{theorem} \label{Theorem:Appendix:GradientEquivalence}
Given a data set $\B$ of $N$ items, loss functions $\Loss_1$ and $\Loss_2$, and priority scheme $pr$, the expected gradient of $\Loss_1(\delta(i))$, where $i \in \B$ is sampled uniformly, is equal to the expected gradient of $\Loss_2(\delta(i))$, where $i$ is sampled with priority $pr$, if $\g_Q \Loss_1(\delta(i)) = \frac{1}{\lambda} pr(i) \g_Q \Loss_2(\delta(i))$ for all $i$, where~$\lambda = \frac{\sum_j pr(j)}{N}$.
\end{theorem}

\textit{Proof.} \begin{equation}
\begin{aligned}
\E_{i \sim \B} \lb \g_Q \Loss_1(\delta(i)) \rb
&= \frac{1}{N} \sum_i \g_Q \Loss_1(\delta(i)) \\
&= \frac{1}{N} \sum_i \frac{N}{\sum_j pr(j)} pr(i) \g_Q \Loss_2(\delta(i)) \\
&= \sum_i \frac{pr(i)}{\sum_j pr(j)} \g_Q \Loss_2(\delta(i)) \\
&= \E_{i \sim pr} \lb \g_Q \Loss_2(\delta(i)) \rb.
\end{aligned}
\end{equation}

\hfill $\blacksquare$ \bigskip

\begin{corollary} \label{Corollary:Appendix:1}
\autoref{Theorem:Appendix:GradientEquivalence} is satisfied by any two loss functions $\Loss_1$, where $i \in \B$ is sampled uniformly, and $\Loss_2$, where $i$ is sampled with respect to priority $pr$, if $\Loss_1(\delta(i)) = \frac{1}{\lambda} |pr(i)|_\times \Loss_2(\delta(i))$ for all $i$, where $\lambda = \frac{\sum_j pr(j)}{N}$ and $|\cdot|_\times$ is the stop-gradient operation.
\end{corollary}

\textit{Proof.} 
\begin{equation}
\begin{aligned}
\g_Q \Loss_1(\delta(i)) &= \g_Q \frac{1}{\lambda} |pr(i)|_\times \Loss_2(\delta(i)) \\
&= \frac{1}{\lambda} pr(i) \g_Q \Loss_2(\delta(i)).
\end{aligned}
\end{equation}
Then $\Loss_1$ and $\Loss_2$ have the same expected gradient by \autoref{Theorem:Appendix:GradientEquivalence}.

\hfill $\blacksquare$ \bigskip

\begin{corollary} \label{Corollary:Appendix:uniformly}
\autoref{Theorem:Appendix:GradientEquivalence} is satisfied by any two loss functions $\Loss_1$, where $i \in \B$ is sampled uniformly, and $\lambda \Loss_2$, where $i$ is sampled with respect to priority $pr$ and $\lambda = \frac{\sum_j pr(j)}{N}$, if $\textnormal{sign}(\g_Q \Loss_1(\delta(i))) = \textnormal{sign}(\g_Q \Loss_2(\delta(i)))$ and $pr(i) = \frac{\g_Q \Loss_1(\delta(i))}{\g_Q \Loss_2(\delta(i))}$ for all $i$.
\end{corollary}

\textit{Proof.} Given $\text{sign}(\g_Q \Loss_1(\delta(i))) = \text{sign}(\g_Q \Loss_2(\delta(i)))$, we have $\text{sign}(pr(i)) = 1$, as we cannot sample with negative priority. \autoref{Theorem:Appendix:GradientEquivalence} is satisfied as:
\begin{equation}
\begin{aligned}
 \frac{1}{\lambda} pr(i) \g_Q \lambda \Loss_2(\delta(i)) &= \frac{\lambda}{\lambda} \cdot \frac{\g_Q \Loss_1(\delta(i))}{\g_Q \Loss_2(\delta(i))} \g_Q \Loss_2(\delta(i)) \\
&= \g_Q \Loss_1(\delta(i)).
\end{aligned}
\end{equation} 

\hfill $\blacksquare$

\subsection{Theorem 2}

\begin{theorem} \label{Theorem:Appendix:Variance}
Given a data set $\B$ of $N$ items and loss function $\Loss_1$, consider the loss function $\lambda \Loss_2(\delta(i))$, where $i \in \B$ is sampled with priority $pr$ and $\lambda = \frac{\sum_j pr(j)}{N}$, such that \autoref{Theorem:Appendix:GradientEquivalence} is satisfied. The variance of $\g_Q \lambda \Loss_2(\delta(i))$ is minimized when $\Loss_2 = \Loss_\textnormal{L1}$ and $pr(i) = |\g_Q \Loss_1(\delta(i))|$.
\end{theorem}

\textit{Proof.}

Consider the variance of the gradient with prioritized sampling. Note $\text{Var}(x) = \E[x^2] - \E[x]^2$.
\begin{equation} \label{eqn:variance}
\begin{aligned} 
\text{Var} \lp \g_Q \lambda \Loss_2(\delta(i)) \rp &= \E_{i \sim pr} \lb \lp \g_Q \lambda \Loss_2(\delta(i)) \rp^2 \rb - \E_{i \sim pr} \lb \g_Q \lambda \Loss_2(\delta(i)) \rb^2\\ 
&= \sum_i \frac{pr(i)}{\sum_j pr(j)} \frac{\lp \sum_j pr(j) \rp^2}{N^2} \lp \g_Q \Loss_2(\delta(i)) \rp^2 - X\\
&= \frac{\sum_j pr(j)}{N^2} \sum_i \g_Q \Loss_1(\delta(i)) \g_Q \Loss_2(\delta(i)) - X, 
\end{aligned}
\end{equation}
where we define $X = \E_{i \sim \B} \lb \g_Q \Loss_1(\delta(i)) \rb^2 = \E_{i \sim pr} \lb \lambda \g_Q \Loss_2(\delta(i)) \rb^2$, the square of the unbiased expected gradient. 

For L1 loss, noting $\text{sign}(\g_Q \Loss_1(\delta(i))) = \text{sign}(\g_Q \Loss_2(\delta(i)))$, then setting $\Loss_2 = \Loss_{\text{L1}}$, we have $pr(i) = |\g_Q \Loss_1(\delta(i))|$ and $\g_Q \Loss_1(\delta(i)) \g_Q \Loss_2(\delta(i)) = |\g_Q \Loss_1(\delta(i))|$, and we can simplify the expression:
\begin{equation} \label{eqn:L1Variance}
\begin{aligned}
&= \frac{\sum_j |\g_Q \Loss_1(\delta(j))|}{N^2} \sum_i \g_Q \Loss_1(\delta(i)) - X \\
&= \lp \frac{\sum_j |\g_Q \Loss_1(\delta(j))|}{N} \rp^2 - X.
\end{aligned}
\end{equation}

Now consider a generic prioritization scheme where $\g_Q \Loss_2(\delta(i)) = f(\delta(i))$. To give the same expected gradient, by Theorem 1 we must have $pr(i) = \g_Q \Loss_1(\delta(i)) / f(\delta(i))$. %
To compute the variance, we can insert these terms into \autoref{eqn:variance}:
\begin{equation}
\begin{aligned}
&=\frac{\sum_j pr(j)}{N^2} \sum_i \g_Q \Loss_1(\delta(i)) \g_Q \Loss_2(\delta(i)) - X \\
&=\frac{\sum_j \g_Q \Loss_1(\delta(j))/f(\delta(j))}{N^2} \sum_i \g_Q \Loss_1(\delta(i)) f(\delta(i)) - X.
\end{aligned}
\end{equation}
Then choosing $u_j = \frac{\sqrt{\g_Q \Loss_1(\delta(j))/f(\delta(j))}}{\sqrt{N}}$ and $v_j = \frac{\sqrt {\g_Q \Loss_1(\delta(j)) f(\delta(j))}}{\sqrt{N}}$, by Cauchy-Schwarz we have:
\begin{equation}
\lp \frac{\sum_j |\g_Q \Loss_1(\delta(j))|}{N} \rp^2 \leq \frac{\sum_j \g_Q \Loss_1(\delta(j))/f(\delta(j))}{N^2} \sum_i \g_Q \Loss_1(\delta(i)) f(\delta(i)),
\end{equation}
with equality if $f(\delta(j)) = \pm c$, where $c$ is a constant. %

It follows that the variance is minimized when $\Loss_2$ is the L1 loss. 

\hfill $\blacksquare$ \bigskip

\begin{observation}
Given a data set $\B$ of $N$ items and loss function $\Loss_1$, the gradient of the loss function $\lambda \Loss_\textnormal{L1}(\delta(i))$, where $i \in \B$ is sampled with priority $pr(i) = |\g_Q \Loss_1(\delta(i))|$ and $\lambda = \frac{\sum_j pr(j)}{N}$, will have lower (or equal) variance than the gradient of $\Loss_1(\delta(i))$, where $i$ is sampled uniformly.
\end{observation}

\textit{Proof.}

This is a direct result of \autoref{Theorem:Appendix:Variance}, by noting setting $\Loss_2 = \Loss_1$, $pr(i) = \frac{1}{N}$ and $\lambda = \frac{\sum_j pr(j)}{N} = 1$. However, to be comprehensive, consider the variance of $\Loss_1$ with uniform sampling. 

\begin{equation}
\begin{aligned}
\text{Var} \lp \g_Q \Loss_1(\delta(i)) \rp
&= \E_{i \sim \B} \lb \lp \g_Q \Loss_1(\delta(i)) \rp^2 \rb - \E_{i \sim \B} \lb \g_Q \Loss_1(\delta(i)) \rb^2\\ 
&= \frac{1}{N} \sum_i \lp \g_Q \Loss_1(\delta(i)) \rp^2 - X.
\end{aligned}
\end{equation}
where $X$ is defined as before. 

Now by the Cauchy-Schwarz inequality $\lp \sum_j u_j v_j \rp^2 \leq \sum_{j=1}^N u_j^2 \sum_{j=1}^N v_j^2$ where $u_j = \frac{1}{\sqrt{N}}$ and $v_j = \frac{|\g_Q \Loss_1(\delta(j))|}{\sqrt{N}}$ we have:
\begin{equation}
\lp \frac{\sum_j |\g_Q \Loss_1(\delta(j))|}{N} \rp^2 \leq \frac{1}{N} \sum_i \lp \g_Q \Loss_1(\delta(i)) \rp^2,
\end{equation}
where the LHS is the variance of L1 loss without the $X$ term, \autoref{eqn:L1Variance}, and so $\text{Var} \lp \g_Q \frac{1}{\lambda} \Loss_2(\delta(i)) \rp$ is less than $\text{Var} \lp \g_Q \Loss_1(\delta(i)) \rp$ for all loss functions $\Loss_1$, when $\Loss_2 = \Loss_\text{L1}$. 

\hfill $\blacksquare$

\subsection{Theorem 3}

\begin{theorem} \label{Theorem:Appendix:PER_tau}
The expected gradient of a loss $\frac{1}{\tau}|\delta(i)|^\tau$, where $\tau > 0$, when used with PER is equal to the expected gradient of the following loss when using a uniformly sampled replay buffer: 
\begin{equation}
\mathcal{L}_\textnormal{PER}^\tau (\delta(i)) = \frac{\eta N }{\tau + \al - \al\beta} |\delta(i)|^{\tau + \al - \al\beta}, \qquad \eta = \frac{\min_j |\delta(j)|^{\al\beta}}{\sum_j |\delta(j)|^\al}.
\end{equation}
\end{theorem}

\textit{Proof.}

For PER, by definition we have $p(i) = \frac{|\delta(i)|^\al}{\sum_{j \in \B} |\delta(j)|^\al}$ and $w(i) = \frac{\lp \frac{1}{N} \cdot \frac{1}{p(i)} \rp^\beta}{\max_{j \in \B} \lp \frac{1}{N} \cdot \frac{1}{p(j)} \rp^\beta}$.

Now consider the expected gradient of $\frac{1}{\tau}|\delta(i)|^\tau$, when used with PER:
\begin{equation}
\begin{aligned}
\E_{i \sim \text{PER}} \lb \g_Q w(i) \frac{1}{\tau}|\delta(i)|^\tau \rb 
&= \sum_{i \in \B} w(i)p(i) \g_Q \frac{1}{\tau}|\delta(i)|^\tau \\
&= \sum_{i \in \B} \frac{\lp \frac{1}{N} \cdot \frac{1}{p(i)} \rp^\beta}{\max_{j \in \B} \lp \frac{1}{N} \cdot \frac{1}{p(j)} \rp^\beta} \frac{|\delta(i)|^\al}{\sum_{j \in \B} |\delta(j)|^\al} \text{sign}(\delta(i)) |\delta(i)|^{\tau - 1} \\
&= \frac{1}{\max_{j \in \B} \frac{1}{|\delta(j)|^{\al\beta}} \sum_{j \in \B} |\delta(j)|^\al} \sum_{i \in \B} \frac{|\delta(i)|^{\tau + \al - 1} \text{sign}(\delta(i))}{|\delta(i)|^{\al\beta}} \\
&= \eta \sum_{i \in \B} \text{sign}(\delta(i)) |\delta(i)|^{\tau + \al - \al\beta - 1}.
\end{aligned}    
\end{equation}

Now consider the expected gradient of $\mathcal{L}_\textnormal{PER}^\tau (\delta(i))$:
\begin{equation}
\begin{aligned}
\E_{i \sim \B} \lb \g_Q \mathcal{L}_\textnormal{PER}^\tau (\delta(i)) \rb 
&= \frac{1}{N }\sum_{i \in \B} \frac{\eta N }{\tau + \al - \al\beta} \g_Q |\delta(i)|^{\tau + \al - \al\beta} \\
&= \eta \sum_{i \in \B} \text{sign}(\delta(i)) |\delta(i)|^{\tau + \al - \al\beta - 1}.
\end{aligned}    
\end{equation}

\hfill $\blacksquare$ \bigskip

\begin{corollary} 
The expected gradient of the Huber loss when used with PER is equal to the expected gradient of the following loss when using a uniformly sampled replay buffer: 
\begin{equation}
    \mathcal{L}_\textnormal{PER}^\textnormal{Huber}(\delta(i)) = \frac{\eta N}{\tau + \al - \al\beta} |\delta(i)|^{\tau + \al - \al\beta}, 
    \qquad \tau =
    \begin{cases}
    2 &\text{if } |\delta(i)| \leq 1,\\
    1 &\text{otherwise,}
    \end{cases}
    \qquad \eta = \frac{\min_j |\delta(j)|^{\al\beta}}{\sum_j |\delta(j)|^\al}.
\end{equation}
\end{corollary}

\textit{Proof.} Direct application of \autoref{Theorem:Appendix:PER_tau} with $\tau=1$ and $\tau=2$.

\hfill $\blacksquare$ \bigskip

\begin{observation}
(MSE) Let $\B(s,a) \subset \B$ be the subset of transitions containing $(s,a)$ and $\delta(i) = Q(i) - y(i)$. If $\g_Q \E_{i \sim \B(s,a)}[0.5 |\delta(i)|^2] = 0$ then $Q(s,a) = \textnormal{mean}_{i \in \B(s,a)} y(i)$.
\end{observation}

\textit{Proof.} 
\begin{equation}
\begin{aligned}
& \E_{i \sim \B(s,a)}[\g_Q 0.5 |\delta(i)|^2] = 0 \\
\Rightarrow~& \E_{i \sim \B(s,a)}[ \delta(i)] = 0 \\
\Rightarrow~& \frac{1}{N} \sum_{i \in \B(s,a)} Q(s,a) - y(i) = 0 \\
\Rightarrow~& Q(s,a) - \frac{2c}{N} \sum_{i \in \B(s,a)} y(i) = 0 \\
\Rightarrow~& Q(s,a) = \frac{1}{N}\sum_{i \in \B(s,a)} y(i).
\end{aligned}
\end{equation}

\hfill $\blacksquare$ \bigskip

\begin{observation}
(L1 Loss) Let $\B(s,a) \subset \B$ be the subset of transitions containing $(s,a)$ and $\delta(i) = Q(i) - y(i)$. If $\g_Q \E_{i \sim \B}[|\delta(i)|] = 0$ then $Q(s,a) = \textnormal{median}_{i \in \B(s,a)} y(i)$.
\end{observation}

\textit{Proof.} 
\begin{equation}
\begin{aligned}
& \E_{i \sim \B(s,a)}[\g_Q |\delta(i)|] = 0 \\
\Rightarrow~& \E_{i \sim \B(s,a)}[\textnormal{sign}(\delta(i))] = 0 \\
\Rightarrow~& \sum_{i \in \B(s,a)} \mathbbm{1}\{Q(s,a) \leq y(i)\} = \sum_{i \in \B(s,a)} \mathbbm{1}\{Q(s,a) \geq y(i)\} \\
\Rightarrow~& Q(s,a) = \textnormal{median}_{i \in \B(s,a)} y(i).
\end{aligned}
\end{equation}

\hfill $\blacksquare$ %

\subsection{PAL Derivation}

\begin{observation}
LAP and PAL have the same expected gradient. 
\end{observation}

\textit{Proof.} From \autoref{Corollary:Appendix:1} we have:

\begin{equation}
\begin{aligned}
\Loss_\text{PAL}(\delta(i))~&= \frac{1}{\lambda} |pr(i)|_\times \Loss_\text{Huber}(\delta(i)) \\
&= \frac{1}{\lambda} |\max(|\delta(i)|^\al,1)|_\times \Loss_\text{Huber}(\delta(i)) \\
&= \frac{1}{\lambda} |\max(|\delta(i)|^\al,1)|_\times 
\begin{cases}
0.5 \delta(i)^2 &\text{if } |\delta(i)| \leq 1,\\
|\delta(i)| &\text{otherwise,}
\end{cases} \\
&= \frac{1}{\lambda} \begin{cases}
0.5 \delta(i)^2 &\text{if } |\delta(i)| \leq 1,\\
\frac{|\delta(i)|^{1 + \al}}{1 + \al} &\text{otherwise,}
\end{cases}
\end{aligned}
\end{equation}
where 
\begin{equation}
\lambda = \frac{\sum_j pr(j)}{N} = \frac{\sum_j \max(|\delta(j)|^\al, 1)}{N}.
\end{equation}
Then by \autoref{Corollary:Appendix:1}, LAP and PAL have the same expected gradient. 

\hfill $\blacksquare$

\section{Computational Complexity Results}

A bottleneck in the usage of prioritized experience replay (PER) \cite{PrioritizedExpReplay} is the computational cost induced by non-uniform sampling. Most implementations of PER use a sum-tree to keep the sampling cost to $O(\log{(N)})$, where $N$ is the number of elements in the buffer. However, we found common implementations to have inefficient aspects, mainly unnecessary for-loops. Since a mini-batch is sampled every time step, any inefficiency can add significant costs to the run time of the algorithm. While our implementation has no algorithmic differences and still relies on a sum-tree, we found it significantly outperformed previous implementations. 

\begin{figure}[ht]
    \centering
    \includegraphics[width=0.4\linewidth]{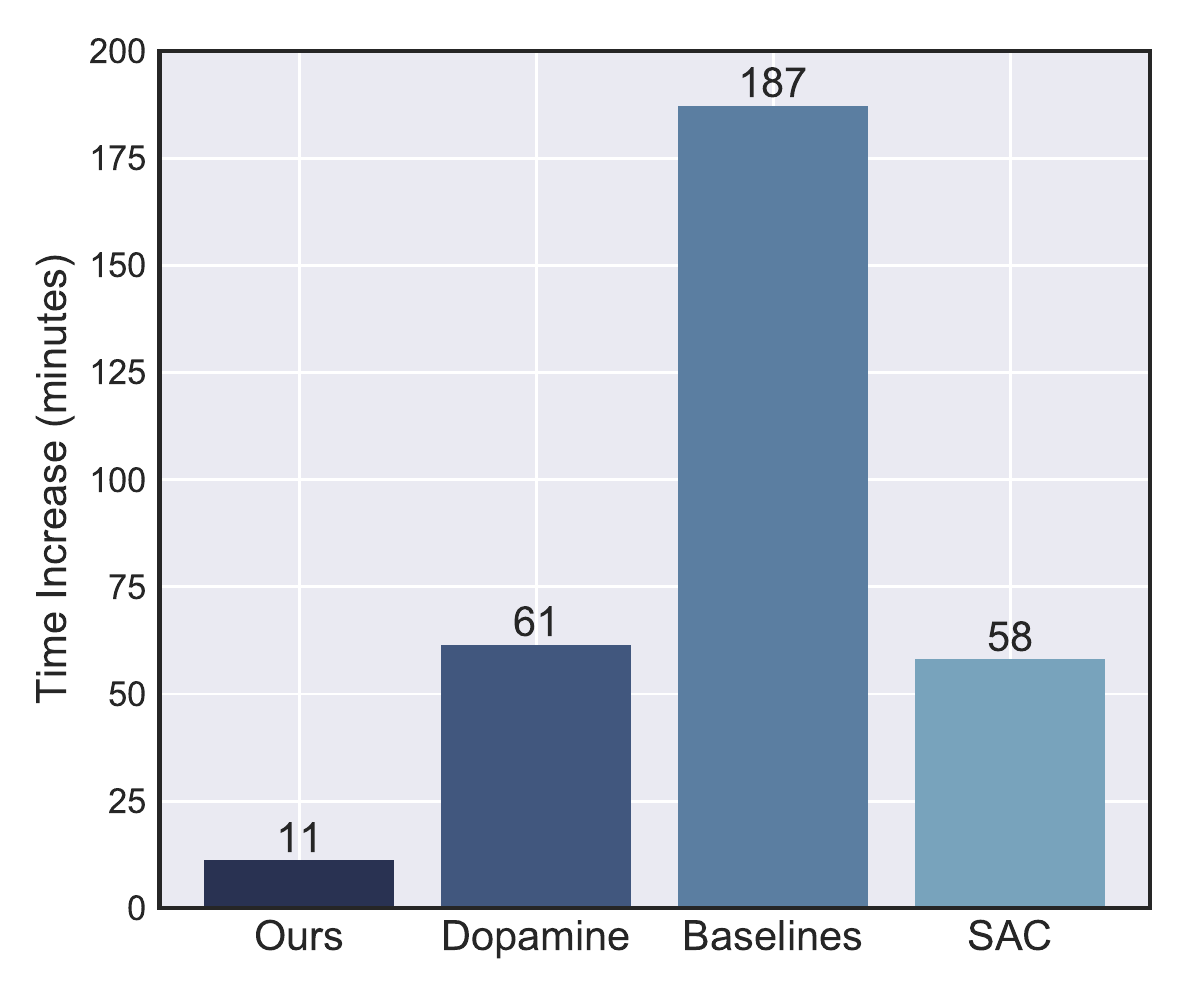}
    \vspace{-3mm}
    \caption{The average run time increase of different implementations of PER in minutes, over 1 million time steps and averaged over 3 trials. Time increase of SAC is provided to give a better understanding of the significance. Our implementation only adds a cost of 11 minutes, per million time steps, while the OpenAI baselines implementation adds over 3 hours.} \label{figure:time}
\end{figure}

We compare the run time of our implementation of PER with two standard implementations, OpenAI baselines \cite{baselines} and Dopamine \cite{castro2018dopamine}. To keep things fair, all components of the experience replay buffer and algorithm are fixed across all comparisons, and only the sampling of the indices of stored transitions and the computation of the importance sampling weights is replaced. Both implementations are taken from the master branch in early February~2020\footnote{OpenAI baselines \url{https://github.com/openai/baselines}, commit: ea25b9e8b234e6ee1bca43083f8f3cf974143998, Dopamine \url{https://github.com/google/dopamine}, commit: e7d780d7c80954b7c396d984325002d60557f7d1.}. Each implementation of PER is combined with TD3. The OpenAI baselines implementation uses an additional sum-tree to compute the minimum over the entire replay buffer to compute the importance sampling weights. However, to keep the computational costs comparable, we remove this additional sum-tree and use a per-batch minimum, similar to Dopamine and our own implementation. Additionally, we compare against TD3 with a uniform experience replay as well as SAC \cite{haarnoja2018soft}. All time-based experiments are run on a single GeForce GTX 1080 GPU and a Intel Core i7-6700K CPU. Our results are presented in~\autoref{figure:time}~and~\autoref{table:time}. 

\begin{table*}[ht]
\centering
\caption{Average run time of different implementations of PER, and their percentage increase over TD3 with a uniform buffer. Values are computed over 1 million time steps and averaged over 3 trials. Run time of TD3 and SAC with uniform buffers are also provided to give a better understanding of the scale. $\pm$ captures a 95\% confidence interval over the run time. Fastest run time implementation of PER is bolded.} \label{table:time}
\small
\begin{tabular}{lccccc}
\toprule
& TD3 + Uniform                   & TD3 + Ours               & TD3 + Dopamine         & TD3 + Baselines             & SAC \\
\midrule
Run Time (mins) & 81.62 $\pm$ 1.14 & \textbf{92.78 $\pm$ 0.93} & 143.02 $\pm$ 1.75 & 268.80 $\pm$ 7.39 & 139.70 $\pm$ 2.22 \\
Time Increase (\%) & +0.00\% & \textbf{+13.68\%} & +75.24\% & +229.35\% & +71.17\% \\ 
\bottomrule
\end{tabular}
\end{table*}

We find our implementation of PER greatly outperforms the other standard implementations in terms of run time. This means PER can be added to most methods without significant computational costs if implemented efficiently. Additionally, we find that TD3 with PER can be run faster than a comparable and commonly used method, SAC. Our implementation of PER adds less than 50 lines of code to the standard experience replay buffer code. We hope the additional efficiency will enable further research in non-uniform sampling methods. 

\section{Additional Experiments}

In this section we perform additional experiments and visualizations, covering ablation studies, additional baselines and display the learning curves for the Atari results. 

\subsection{Ablation Study}

To better understand the contributions of each component in PAL, we perform an ablation study. We aim to understand the importance of the scaling factor $\frac{1}{\lambda} = \frac{N}{\sum_j \max(|\delta(j)|^\al, 1)}$ as well as the differences between the proposed loss and the comparable Huber loss \cite{huber1964robust}, by considering all possible combinations. Notably, when $\al=0$, PAL equals the Huber loss with the scaling factor. As discussed in the Experimental Details, \autoref{appendix:experimental_details}, PAL uses $\al=0.4$. The results are reported in \autoref{appendix:figure:ablation} and \autoref{appendix:table:ablation}. 

We find the complete loss of PAL achieves the highest performance, and PAL without the $\frac{1}{\lambda}$ scale factor to be the second highest. Interestingly, while TD3 with the Huber loss performs poorly, scaling by $\frac{1}{\lambda}$ adds fairly significant gains in performance.

\begin{figure*}[ht]
    \centering
    \includegraphics[width=\linewidth]{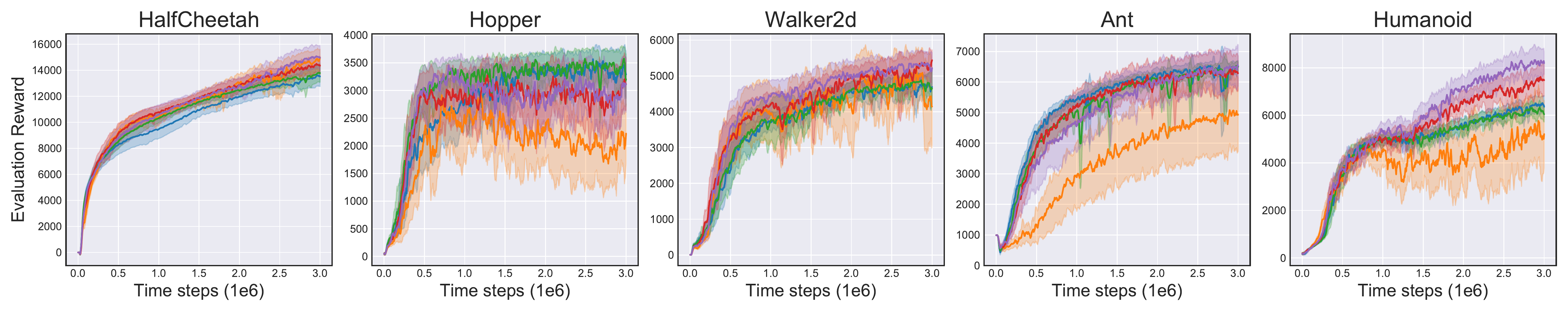}
    \includegraphics[scale=\thelastscalefactor]{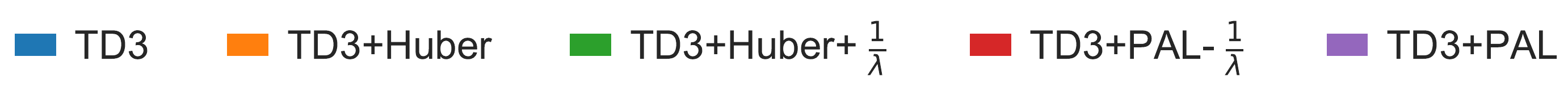}
    \caption{Learning curves for the ablation study on the suite of OpenAI gym continuous control tasks in MuJoCo. Curves are averaged over 10 trials, where the shaded area represents a 95\% confidence interval over the trials.} \label{appendix:figure:ablation}%
    \vskip -0.1in
\end{figure*}

\setlength{\tabcolsep}{3.4pt}
\begin{table}[ht]
\centering
\caption{Average performance over the last 10 evaluations and 10 trials. $\pm$ captures a 95\% confidence interval. Scores are bold if the confidence interval intersects with the confidence interval of the highest performance, except for HalfCheetah and Walker2d where all scores satisfy this condition.} \label{appendix:table:ablation}
\small
\begin{tabular}{lccccc}
\toprule
& TD3 & TD3 + Huber & TD3 + Huber + $\frac{1}{\lambda}$ & TD3 + PAL - $\frac{1}{\lambda}$ & TD3 + PAL \\
\midrule
HalfCheetah & 13570.9 $\pm$ 794.2 & 14820.5 $\pm$ 785.5 & 13772.2 $\pm$ 685.7 & 14404 $\pm$ 642.2 & 15012.2 $\pm$ 885.4 \\
Hopper      & \textbf{3393.2 $\pm$ 381.9} & 2125.7 $\pm$ 596.9 & \textbf{3442.2 $\pm$ 319.4} & \textbf{3135.8 $\pm$ 479.5} & \textbf{3129.1 $\pm$ 473.5} \\
Walker2d    & 4692.4 $\pm$ 423.6 & 4311.3 $\pm$ 1219.2 & 4707.3 $\pm$ 435.3 & 5313.7 $\pm$ 368.2 & 5218.7 $\pm$ 422.6 \\
Ant         & \textbf{6469.9 $\pm$ 200.3} & 4952.6 $\pm$ 1204.2 & \textbf{6499.2 $\pm$ 162.7} & \textbf{6322.7 $\pm$ 564.3} & \textbf{6476.2 $\pm$ 640.2} \\
Humanoid    & 6437.5 $\pm$ 349.3 & 5039.1 $\pm$ 1631.5 & 6163.2 $\pm$ 331.7 & \textbf{7493.4 $\pm$ 645.3} & \textbf{8265.9 $\pm$ 519.0} \\
\bottomrule
\end{tabular}
\end{table}

\subsection{Additional Baselines}

To better compare our algorithm against other recent adjustments to replay buffers and non-uniform sampling, we compare LAP and PAL against the Emphasizing Recent Experience (ERE) replay buffer \cite{wang2019towards} combined with TD3. To implement ERE we modify our training procedure slightly, such that $X$ training iterations are the applied at the end of each episode, where $X$ is the length of the episode. ERE works by limiting the transitions sampled to only the $N$ most recent transitions, where $N$ changes over the $X$ training iterations. Additionally, we add a baseline of the simplest version of LAP which uses the L1 loss, rather than the Huber loss and samples transitions with priority $pr(i)=|\delta(i)|^\al$, denoted TD3+L1+$\al$. Results are reported in \autoref{appendix:figure:baselines} and \autoref{appendix:table:baselines}.

We find that LAP and PAL with TD3 outperform ERE. The addition of ERE outperforms vanilla TD3 and improves early learning performance in several tasks. We remark that the addition of ERE does not directly conflict with LAP and PAL. PAL can be directly combined with ERE with no other modifications. LAP can be combined by only performing the non-uniform sampling on the corresponding subset of transitions determined by ERE. We leave these combinations to future work. We also find that LAP and PAL outperform the simplest version of LAP with the L1 loss. This demonstrates the importance of the Huber loss in LAP. 

\begin{figure*}[ht]
    \centering
    \includegraphics[width=\linewidth]{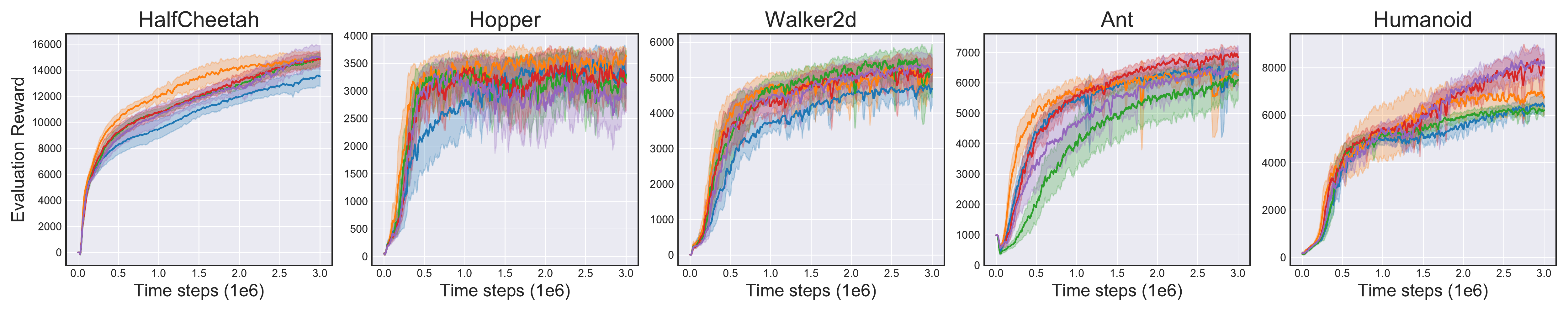}
    \includegraphics[scale=\thelastscalefactor]{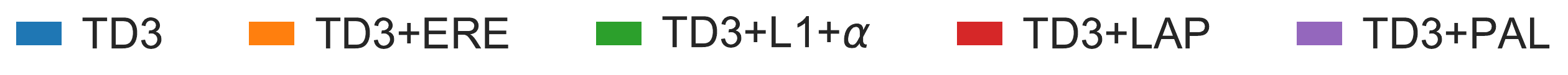}
    \caption{Learning curves for additional baselines on the suite of OpenAI gym continuous control tasks in MuJoCo. Curves are averaged over 10 trials, where the shaded area represents a 95\% confidence interval over the trials.} \label{appendix:figure:baselines}%
    \vskip -0.1in
\end{figure*}

\setlength{\tabcolsep}{3.8pt}
\begin{table}[ht]
\centering
\caption{Average performance over the last 10 evaluations and 10 trials. $\pm$ captures a 95\% confidence interval. Scores are bold if the confidence interval intersects with the confidence interval of the highest performance, except for HalfCheetah, Hopper and Walker2d where all scores satisfy this condition.} \label{appendix:table:baselines}
\small
\begin{tabular}{lccccc}
\toprule
            & TD3                   & TD3 + ERE               & TD3 + L1 + $\al$         & TD3 + LAP             & TD3 + PAL \\
\midrule
HalfCheetah & 13570.9 $\pm$ 794.2 & 14863.6 $\pm$ 602.0 & 14885.4 $\pm$ 402.6 & 14836.5 $\pm$ 532.2 & 15012.2 $\pm$ 885.4 \\
Hopper      & 3393.2 $\pm$ 381.9 & 3541.7 $\pm$ 253.3 & 3208.9 $\pm$ 475.1 & 3246.9 $\pm$ 463.4 & 3129.1 $\pm$ 473.5 \\
Walker2d    & 4692.4 $\pm$ 423.6 & 5205.5 $\pm$ 407.0 & 5153.7 $\pm$ 601.4 & 5230.5 $\pm$ 368.2 & 5218.7 $\pm$ 422.6 \\
Ant         & 6469.9 $\pm$ 200.3 & 6285.9 $\pm$ 271.7 & 6021.1 $\pm$ 656.9 & \textbf{6912.6 $\pm$ 234.4} & \textbf{6476.2 $\pm$ 640.2} \\
Humanoid    & 6437.5 $\pm$ 349.3 & 6889.9 $\pm$ 627.4 & 6185.3 $\pm$ 207.0 & \textbf{7855.6 $\pm$ 705.9} & \textbf{8265.9 $\pm$ 519.0} \\
\bottomrule
\end{tabular}
\end{table}

\subsection{Full Atari Results}

We display the full learning curves from the Atari experiments in \autoref{appendix:figure:atari}. 

\begin{figure*}[ht]
    \centering
    \includegraphics[width=\linewidth]{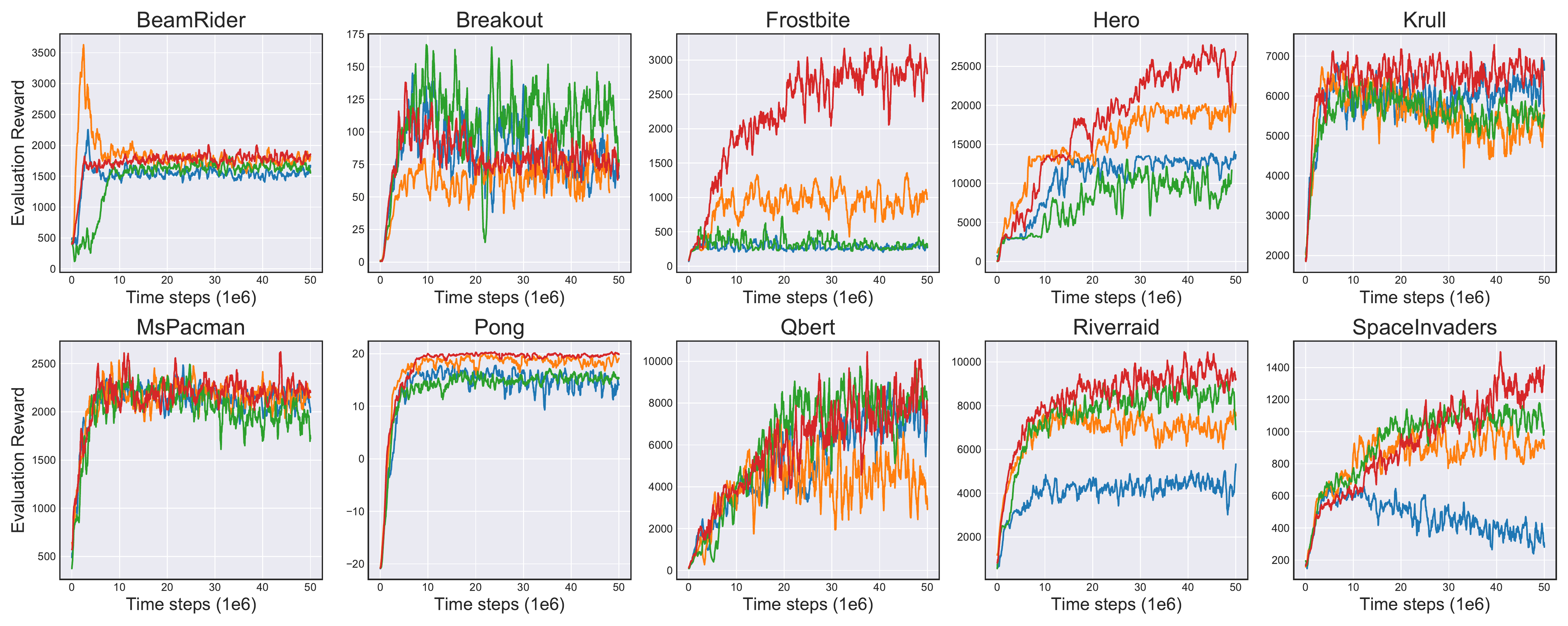}
    \includegraphics[scale=\thelastscalefactor]{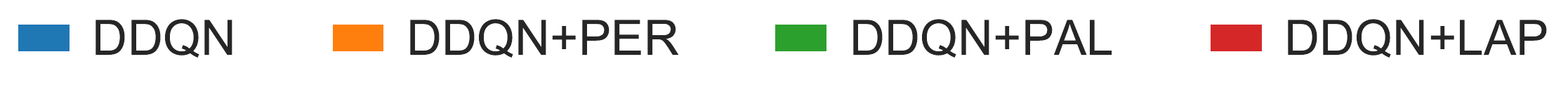}
   \caption{Complete learning curves for a set of Atari games. Curves are smoothed uniformly with a sliding window of $10$ for visual clarity.} \label{appendix:figure:atari}%
    \vskip -0.1in
\end{figure*}

\clearpage

\section{Experimental Details} \label{appendix:experimental_details}

All networks are trained with PyTorch (version 1.2.0) \cite{paszke2019pytorch}, using PyTorch defaults for all unmentioned hyper-parameters.

\subsection{MuJoCo Experimental Details}

\textbf{Environment.} Our agents are evaluated in MuJoCo (mujoco-py version 2.0.2.9) \cite{mujoco} via OpenAI gym (version 0.15.4) \cite{OpenAIGym} interface, using the v3 environments. The environment, state space, action space, and reward function are not modified or pre-processed in any way, for easy reproducibility and fair comparison with previous results. Each environment runs for a maximum of $1000$ time steps or until some termination condition, and has a multi-dimensional action space with values in the range of $(-1,1)$, except for Humanoid which uses a range of $(-0.4, 0.4)$.

\textbf{Architecture.} Both TD3 \cite{fujimoto2018addressing} and SAC \cite{haarnoja2018soft} are actor-critic methods which use two Q-networks and a single actor network. All networks have two hidden layers of size $256$, with ReLU activation functions after each hidden layer. The critic networks take state-action pairs $(s,a)$ as input, and output a scalar value $Q$ following a final linear layer. The actor network takes state $s$ as input and outputs a multi-dimensional action $a$ following a linear layer with a tanh activation function, multiplied by the scale of the action space. For clarity, a network definition is provided in \autoref{fig:network}.

\begin{figure}[ht]
\centering
\begin{BVerbatim}
(input dimension, 256)
ReLU
(256, 256)
RelU
(256, output dimension)
\end{BVerbatim}
\caption{Network architecture. Actor networks are followed by a \texttt{tanh} $\cdot$ \texttt{max action size}} \label{fig:network}
\end{figure}

\textbf{Network Hyper-parameters.} Networks are trained with the Adam optimizer \cite{adam}, with a learning rate of $3 \text{e} -4$ and mini-batch size of $256$. The target networks in both TD3 and SAC are updated with polyak averaging with $\nu=0.005$ after each learning update, such that $\ta' \leftarrow (1 - \nu) \ta' + \nu \ta$, as described by \citet{DDPG}. 

\textbf{Terminal Transitions.} The learning target uses a discount factor of $\y=0.99$ for non-terminal transitions and $\y=0$ for terminal transitions, where a transition is considered terminal only if the environment ends due to a termination condition and not due to reaching a time-limit. 

\textbf{LAP, PAL, and PER.} For LAP and PAL we use $\al=0.4$. For PER we use $\al=0.6$, $\beta=0.4$ as described by \citet{PrioritizedExpReplay} and $\e=1 \text{e} -10$ from \citet{castro2018dopamine}. Since there are two TD errors defined by $\delta_1 = Q^1_\ta - y$ and $\delta_2 = Q^2_\ta - y$, where $y = r + \y \min(Q^1_{\ta'}(s',a'), Q^2_{\ta'}(s',a'))$ \cite{fujimoto2018addressing}, the priority uses the maximum over $|\delta_1|$ and $|\delta_2|$, which we found to give the strongest performance. As done by PER, new samples are given a priority equal to the maximum priority recorded at any point during learning. 

\textbf{TD3 and SAC.} For TD3 we use the default policy noise of $\N(0,\sigma_N^2)$, clipped to $(-0.5, 0.5)$, where $\sigma_N=0.2$. Both values are scaled by the range of the action space. For SAC we use the learned entropy variant \cite{haarnoja2018applications}, where entropy is trained to a target of $- \texttt{action dimensions}$ as described the author, with an Adam optimizer with learning $3 \text{e} -4$, matching the other networks. Following the author's implementation, we clip the log standard deviation to $(-20, 2)$, and add $\e=1 \text{e} -6$ to avoid numerical instability in the logarithm operation. We consider this a hyper-parameter as $\log{(\e)}$ defines the minimum value subtracted from log-likelihood calculation of the tanh normal distribution:
\begin{equation}
    \log \pi(a|s) = \sum_i -0.5\frac{(u_i - \mu_i)^2}{\sigma_i^2} - \log{\sigma_i} - 0.5 \log{(2 \pi)} - \log{(1 - a_i^2 + \e)},
\end{equation}
where $i$ is each action dimension, $u$ is pre-activation value of the action $a$ and $\mu$ and $\sigma$ define the Normal distribution output by the actor network. The log-likelihood calculation assumes the action is scaled within $(-1,1)$. As standard practice, the agents are trained for one update after every time step.

\textbf{Exploration.} To fill the buffer, for the first $25$k time steps the agent is not trained, and actions are selected randomly with uniform probability. Afterwards, exploration occurs in TD3 by adding Gaussian noise $\N(0,\sigma_E^2 \cdot \texttt{max action size})$, where $\sigma_E = 0.1$ scaled by the range of the action space. No exploration noise is added to SAC, as it uses a stochastic policy. 

For clarity, all hyper-parameters are presented in \autoref{table:hyperparameters}.

\begin{table}[ht]
\centering
\caption{Continuous control hyper-parameters.} \label{table:hyperparameters}
\begin{center}
\begin{tabular}{lc}
\toprule
Hyper-parameter & Value \\
\midrule
Optimizer & Adam \\
Learning rate & $3 \text{e} -4$ \\
Mini-batch size & $256$ \\
Discount factor $\y$ & $0.99$ \\
Target update rate & $0.005$ \\
Initial random policy steps & $25$k \\
\midrule
TD3 Exploration Policy $\sigma_E$ & $0.1$ \\
TD3 Policy Noise $\sigma_N$ & $0.2$ \\
TD3 Policy Noise Clipping & $(-0.5, 0.5)$ \\
\midrule
SAC Entropy Target & $- \texttt{action dimensions}$ \\
SAC Log Standard Deviation Clipping & $(-20, 2)$ \\
SAC $\e$ & $1 \text{e} -6$ \\
\midrule
PER priority exponent $\al$ & $0.6$ \\
PER importance sampling exponent $\beta$ & $0.4$ \\
PER added priority $\e$ & 1 $\text{e} -10$ \\
LAP \& PAL exponent $\al$ & $0.4$ \\
\bottomrule
\end{tabular}
\end{center}
\end{table}

\textbf{Hyper-parameter Optimization.} No hyper-parameter optimization was performed on TD3 or SAC. For LAP and PAL we tested $\al=\{0.2, 0.4, 1\}$ on the HalfCheetah task using seeds than the final results reported. Note $0.4$ comes from $\al-\al\beta=0.36$ when using the hyper-parameters defined by \citet{PrioritizedExpReplay}. Additionally we tested using defining the priority over $|\delta_1|$, $0.5(|\delta_1| + |\delta_2|)$ and $\max(|\delta_1|,|\delta_2|)$ and found the maximum to work the best for both LAP and PER. 

\textbf{Evaluation.} Evaluations occur every $5000$ time steps, where an evaluation is the average reward over $10$ episodes, following the deterministic policy from TD3, without exploration noise, or by taking the deterministic mean action from SAC. To reduce variance induced by varying seeds \cite{hendersonRL2017}, we use a new environment with a fixed seed (the training seed + a constant) for each evaluation, so each evaluation uses the same set of initial start states.

\textbf{Visualization.} Performance is displayed in learning curves, presented as the average over $10$ trials, where a shaded region is added, representing a $95\%$ confidence interval over the trials. The curves are smoothed uniformly over a sliding window of $5$ evaluations for visual clarity. 

\subsection{Atari Experimental Details}

\textbf{Environment.} For the Atari games we interface through OpenAI gym (version 0.15.4) \cite{OpenAIGym} using NoFrameSkip-v0 environments with sticky actions with $p=0.25$ \cite{machado2018revisiting}. Our pre-processing steps follow standard procedures as defined by \citet{castro2018dopamine} and \citet{machado2018revisiting}. Exact pre-processing steps can be found in \cite{fujimoto2019benchmarking} which our code is closely based on. 

\textbf{Architecture.} We use the standard architecture used by DQN-style algorithms. This is a 3-layer CNN followed by a fully connected network with a single hidden layer. The input to this network is a $(4,84,84)$ tensor, corresponding to the previous $4$ states of $84 \times 84$ pixels. The first convolutional layer has $32$ $8 \times 8$ filers with stride $4$, the second convolutional layer has $32$ $4 \times 4$ filers with stride $2$, the final convolutional layer has $64$ $2 \times 2$ filers with stride $1$. The output is flattened into a $3136$ vector and passed to the fully connected network with a single hidden layer of $512$. The final output dimension is the number of actions. All layers, except for the final layer are followed by a ReLU activation function. 

\textbf{Network Hyper-parameters.} Networks are trained with the RMSProp optimizer, with a learning rate of $6.25 \text{e} -5$, no momentum, centered, $\al=0.95$, and $\e=1$e$-5$. We use a mini-batch size of $32$. The target network is updated every $8$k training iterations. A training update occurs every $4$ time steps (i.e. $4$ interactions with the environment). 

\textbf{Terminal Transitions.} The learning target uses a discount factor of $\y=0.99$ for non-terminal transitions and $\y=0$ for terminal transitions, where a transition is considered terminal only if the environment ends due to a termination condition and not due to reaching a time-limit. 

\textbf{LAP, PAL, and PER.} For LAP and PAL we use $\al=0.6$. For PER we use $\al=0.6$, $\beta=0.4$ as described by \citet{PrioritizedExpReplay} and $\e=1 \text{e} -10$ from \citet{castro2018dopamine}. As done by PER, new samples are given a priority equal to the maximum priority recorded at any point during learning. For LAP, due to the magnitude of training errors being lower, rather than use a Huber loss with $\kappa=1$, we use $\kappa=0.01$. This gives a loss function of:
\begin{equation}
    \mathcal{L}_\text{Huber}^{\kappa}(\delta(i)) = 
    \begin{cases}
    0.5 \delta(i)^2 &\text{if } |\delta(i)| \leq \kappa,\\
    \kappa (|\delta(i)| - 0.5\kappa) &\text{otherwise.}
    \end{cases}
\end{equation}
To maintain the unbiased nature of LAP, the priority clipping needs to be shifted according to the choice of $\kappa$, generalizing the priority function to be:
\begin{equation}
p(i) = \frac{\max(|\delta(i)|^\al,\kappa^\al)}{\sum_j \max(|\delta(j)|^\al,\kappa^\al)}.
\end{equation}
Note the case where $\kappa=1$, the algorithm is unchanged. PAL also needs to be generalized accordingly:
\begin{equation}
    \mathcal{L}_{\text{PAL}}(\delta(i)) = 
    \frac{1}{\lambda} \begin{cases}
    0.5 \kappa^\al \delta(i)^2 &\text{if } |\delta(i)| \leq \kappa,\\
    \frac{\kappa |\delta(i)|^{1 + \al}}{1 + \al} &\text{otherwise,}
    \end{cases} \qquad \lambda = \frac{\sum_j \max(|\delta(j)|^\al, \kappa^\al)}{N}.
\end{equation}

\textbf{Exploration.} To fill the buffer, for the first $20$k time steps the agent is not trained, and actions are selected randomly with uniform probability. Afterwards, exploration occurs according to the $\e$-greedy policy where $\e$ begins at $1$ and is decayed to $0.01$ over the next $250$k training iterations, corresponding to $1$ million time steps. 

For clarity, all hyper-parameters are presented in \autoref{table:atari_hyperparameters}.

\begin{table}[ht]
\centering
\caption{Atari Hyper-parameters.} \label{table:atari_hyperparameters}
\begin{center}
\begin{tabular}{lc}
\toprule
Hyper-parameter & Value \\
\midrule
Optimizer & RMSProp \\
RMSProp momentum & $0$ \\ 
RMSProp centered & True \\
RMSProp $\al$ & $0.95$ \\
RMSProp $\e$ & $1 \text{e} -5$ \\
Learning rate & $6.25 \text{e} -5$ \\
Mini-batch size & $32$ \\
Discount factor $\y$ & $0.99$ \\
Target update rate & $8$k training iterations \\
Initial random policy steps & $20$k \\
Initial $\e$ & $1$ \\
End $\e$ & $0.01$ \\
$\e$ decay period & $250$k training iterations \\ 
Evaluation $\e$ & $0.001$ \\
Training frequency & $4$ time steps \\
\midrule
PER priority exponent $\al$ & $0.6$ \\
PER importance sampling exponent $\beta$ & $0.4$ \\
PER added priority $\e$ & 1 $\text{e} -10$ \\
LAP \& PAL exponent $\al$ & $0.6$ \\
\bottomrule
\end{tabular}
\end{center}
\end{table}

\textbf{Hyper-parameter Optimization.} No hyper-parameter optimization was performed on Double DQN. For LAP and PAL we tested $\al=\{0.4, 0.6\}$ on the Asterix game, which was excluded from the final results.

\textbf{Evaluation.} Evaluations occur every $50$k time steps, where an evaluation is the average reward over $10$ episodes, following the $\e$-greedy policy where $\e=1$e$-3$. To reduce variance induced by varying seeds \cite{hendersonRL2017}, we use a new environment with a fixed seed (the training seed + a constant) for each evaluation, so each evaluation uses the same set of initial start states. Final scores reported in Figure 2 and Table 2 average over the final $10$ evaluations. For Table 2 percentage improvement of $x$ over $y$ is calculated by $\frac{x-y}{|y|}$.

\end{document}